\documentclass{new_tlp}

\usepackage{amsfonts, amsmath, amssymb, mathrsfs}
\usepackage{times, helvet, courier}
\usepackage{graphicx, epsfig, epstopdf}
\usepackage{wrapfig}
\usepackage{multirow}
\usepackage{url}
\usepackage{stmaryrd}
\usepackage[mathscr]{euscript} 
\usepackage{soul}
\usepackage{todonotes}
\usepackage{latexsym}
\usepackage{amssymb}
\usepackage{acronym}
\usepackage[utf8]{inputenc}
\usepackage{microtype}
\usepackage{algorithm}
\usepackage{algorithmic}
\usepackage[algo2e,ruled,linesnumbered,vlined]{algorithm2e}

\usepackage{color}

\def\naf{{ not \:}}

\def\causes{\hbox{\bf causes}}
\def\caused{\hbox{\bf caused}}
\def\executable{\hbox{\bf executable}}
\def\initially{\hbox{\bf initially}}
\def\lan{\langle}
\def\ran{\rangle}
\def\cpasp{{\sc CPasp}}
\def\determines{\hbox{\bf determines}}
\def\oneof{\hbox{\bf oneof}}
\def\kcases{{\bf cases}}
\def\cps{{\sc ASPcp}}

\def\cps{LCP}
\def\cpasp{{\sc CPasp}}

\def\kacmbp{KACMBP}
\def\cpa{{\sc CpA}}
\def\pond{POND}
\def\cff{CFF}
\def\cplan{$\cal C$-{\sc Plan}}

\def\cpa{{\textnormal{\sc CpA}}}

\def\cf2cff{{\tt cf2cs(ff)}}

\def\cpls{{\tt CpLs}}

\def\t0{{\tt t0}}
\def\T1{{\tt t1}}

\def\pond{{\sc POND}}

\def\cff{{\sc CFF}}
\def\kacmbp{{\sc KACMBP}}
\def\dnf{{\sc DNF}}
\def\cnf{{\sc CNF}}

\def\cplan{{\sc C-Plan}}

\def\gclama{{\sc GC}[LAMA]}

\def\dnf{{\sc Dnf}}
\def\cnf{{\sc Cnf}}
\def\pip{{\sc PIP}}

\def\until{\hbox{\bf until}}
\def\always{\hbox{\bf always}}
\def\eventually{\hbox{\bf eventually}}
\def\next{\hbox{\bf next}}

\newcommand{\dlvk}{\dlv$^\mathcal{K}$}

\newcommand{\sysfont}{\textit}

\newcommand{\Clingo}{\sysfont{Clingo}}

\newcommand{\asprin}{\sysfont{asprin}}

\newcommand{\clasp}{\sysfont{clasp}}

\newcommand{\clingo}{\sysfont{clingo}}
\newcommand{\cmodels}{\sysfont{cmodels}}

\newcommand{\dlv}{\sysfont{dlv}}

\newcommand{\plasp}{\sysfont{plasp}}

\newcommand{\smodels}{\sysfont{smodels}}

\newcommand{\clingoM}[1]{\clingo{\small\textnormal{[}\textsc{#1}\textnormal{]}}}

\newtheorem{corollary}{Corollary}
\newtheorem{proposition}{Proposition}

\newtheorem{theorem}{Theorem}
\newtheorem{example}{Example}
\newtheorem{definition}{Definition}
\newtheorem{remark}{Remark}

\usepackage{listings}
\lstdefinelanguage{clingo}{
  keywordstyle=[1]\usefont{OT1}{cmtt}{m}{n},%
  keywordstyle=[2]\textbf,%
  keywordstyle=[3]\usefont{OT1}{cmtt}{m}{n},
  alsoletter={\#,\&},%
  keywords=[1]{not,from,import,exists,if,else,return,while,             break,and,or,for,in,del,and,class,subClass,concern,aspect,subCo,prop,rdf,cpsf,addBy,suppBy,neg,d,relation,holds,h,obs,action,fluent,occurs,req,group,leadTo,active,deg,comp,order,hSubCo,llh_sat_sub,llh_sat,llh_sat_sub_aux,step,deg_pos,nAllPosCon,nActPosCon,possImpactsPos,scoreLoS,wBus,wHum,wTru,wFun,wTim,wBou,wLif,wCom,wDat,last,addFun,func,formula,sat_formula,possImpactsNeg,conflict,member,sa_action,exec,sc_concern},%
  keywords=[2]{\#const,\#show,\#minimize,\#maximize,\#base,\#theory,\#count,\#external,\#program,\#script,\#end,\#heuristic,\#edge,\#project,\#show},%
  keywords=[3]{&,&dom,&sum,&diff,&show,&minimize},%
  morecomment=[l]{\#\ },%
  morecomment=[l]{\%\ },%
  commentstyle={\color{darkgray}}%
}
\allowdisplaybreaks

\begin{document}


\title{Answer Set Planning: A Survey}

\author[T.C.\ Son, E. \ Pontelli, M.\ Balduccini, T.\ Schaub]
{TRAN CAO SON and ENRICO PONTELLI \\
  Department of Computer Science, New Mexico State University\\
  \email{tson | epontell@cs.nmsu.edu}
  \and
  MARCELLO BALDUCCINI\\
  Department of Decision and System Sciences, Saint Joseph's University\\
  \email{mbalducc@sju.edu}
  \and
  TORSTEN SCHAUB\\
  Department of Computer Science, University of Potsdam\\
  \email{torsten@cs.uni-potsdam.de}
}

\maketitle

\begin{abstract}
\emph{Answer Set Planning} refers to the use of \emph{Answer Set Programming (ASP)} to compute \emph{plans}, i.e., solutions to planning problems, that transform a given state of the world to another state. The development of efficient and scalable answer set solvers has provided a significant boost to the development of ASP-based planning systems. This paper surveys the progress made during the last two and a half decades in the area of answer set planning, from its foundations to its use in challenging planning domains. The survey explores the advantages and disadvantages of answer set planning. It also discusses  typical applications of answer set planning and presents a set of challenges for future research. 
\end{abstract}


\section{Introduction}\label{sec:introduction}



Automated planning represents one of the core components in the design of autonomous intelligent systems. 
The term refers to the task of finding a course of actions (i.e., a \emph{plan}) that changes the state of the world from a given state to another state. 
An automated planner takes a planning problem as input, which consists of a domain description or an action theory, the initial state description, and the goal state description, and computes a solution of the problem if one exists.
Automated planning has been an active research area of Artificial Intelligence for many years. It has established itself as a mature research area with its own annually conference, the International Conference on Automated Planning and Scheduling (ICAPS)\footnote{
   \url{https://www.icaps-conference.org}  
} series starting from 1991, with several satellite workshops related to planning and scheduling as well as the planning competition for many tracks.  
Consequently, the literature on planning is enormous. 
The textbooks by \citeN{ghallabnt04} and \citeyear{ghallabnt16} includes more than 500 and 600 references, respectively. The monograph on planning with focus on abstraction and decomposition by \citeN{yang97} has more than 150 references. 
The survey on classical planning by \citeN{hetadr90a} also referred to more than 100 papers. 
Similar observation can be made about the survey by \citeN{weld94a}, which mainly discusses partial order planning.
There are also special collections or special issues on planning such as \cite{allen91,allenht90}. 
In addition, several planning systems addressing different aspects of planning have been developed,\footnote{
   Detailed references to these systems are provided in the subsequent sections. 
} which will be discussed in more details in Sections \ref{asp-sec}--\ref{section:sensing}. 
Our goal in this paper is to provide a survey on answer set planning, a relatively late addition to the rich body of research in automated planning that has not been comprehensively surveyed so far.

\emph{Answer set planning}, a term coined by \citeN{lifschitz99b}, refers to the use of \emph{Answer Set Programming (ASP)} in planning. In this approach, planning problems are translated to logic programs whose answer sets correspond to solutions of the original planning problems. 
This approach is related to the early approach to planning using automated theorem provers by \citeN{green69a}. Although similar in the emphasis of using a general logical solver for planning, answer set planning and planning using automated theorem provers differ in that the former computes an answer set (or a full model) to find a solution whilst the latter identifies a solution with a proof of a query.  
Answer set planning is more closely related to the prominent approach of planning using satisfiability solvers (SAT-planning) proposed by \citeN{kausel92b} and \citeN{kamcse96a}, who showed, experimentally, that SAT-planning can reach the scalability and efficiency of contemporary heuristic-based planning systems. This success is, likely, one of the driving forces behind the research on using logic  programs under the answer set semantics for planning. The idea of answer set planning was first discussed by  \citeN{subzan95a} and further developed by \citeN{dineko97a}, who also demonstrated that answer set planning can be competitive with state-of-the-art domain-independent planner at the time.

Answer set planning offers a number of features which are advantageous for researchers. First, by virtue of the declarative nature of logic programming, answer set planning is itself declarative and elaboration tolerant. 
This enables a modular development of planning systems with special features. For example, 
to consider a particular set of solutions satisfying an user's preferences, one only needs to develop rules expressing these preferences and adds them to the set of rules encoding the planning problem \cite{sonpon06a}; to exploit the various types of domain-knowledge in planning, one only needs to develop rules expressing them to the set of rules encoding the planning problem \cite{sobanamc03a}. To the best of our knowledge, there exists no other planning system that can simultaneously exploit all three well-known types of domain knowledge---temporal, hierarchical, and procedural knowledge---as demonstrated by \cite{sobanamc03a}. 
Other features of logic programming  such as the use of variables, constraints, and choice atoms allow for a compact representation of answer set planners. For example, the basic code for generating a plan in classical setting requires only 10 basic rules and one constraint while a traditional implementation of a planning system in an imperative language  may require thousands of lines.  
Second, the expressiveness of logic programming facilitates the integration of complex reasoning, such as reasoning with static causal laws, into ASP-based planners. To the best of our knowledge, only answer set planning systems deal directly with unrestricted static causal laws \cite{sotugemo05b,tusogemo11a}.  Third, as demonstrated in several experimental evaluations \cite{gekaotroscwa13a,sotugemo05b,tusoba07a,tusogemo11a}, answer set planning systems perform well against other contemporary planning systems in various categories. 
Finally, the large body of theoretical building block results in logic programming supports the development of provably correct answer set planners. This is an important feature of answer set planning that is mostly neglected by the vast majority of work in planning, arguably difficult to obtain for planners realized using traditional imperative programming techniques and valuable for foundational research work. 

Over the last twenty-five years, a variety of ASP-based planning systems have been developed,
e.g. \cite{digelurosc18a,eifalepfpo00a,eifalepfpo03b,gekaotroscwa13a,sotugemo05b,tusoba07a,tusogemo11a,motuso07a,fandinno21,rizwan20,spies19,yalciner17}, that  address several challenges in
planning, such as planning with incomplete information, non-deterministic actions, and sensing actions.
These systems,  in turn, provide the basis for investigation of ASP solutions to problems in areas like diagnosis \cite{balgel03b}, multi-agent path findings \cite{ngobsoscye17a,goheba20a}, goal recognition design \cite{sosascscye16a}, planning with preferences \cite{sonpon06a}, planning with action cost \cite{eifalepfpo03c}, and
robot task planning \cite{jizhkhst19a}.  
This progress has been amplified by the development of efficient and scalable answer set solvers,
 such as \smodels\  \cite{niesim97a}, \dlv{} \cite{dlv97a,alcadofuleperiveza17a,aldofaleri13a}, \clasp\ \cite{gekanesc07b,gekakasc17a}, and \cmodels\ \cite{liemar04a}, together with the invention of action languages for reasoning about actions and change, such as the action languages $\mathcal{A}$,  $\mathcal{B}$, and $\mathcal{C}$ \cite{gellif98a},  $\mathcal{A}_K$ with sensing actions \cite{lometa97a,sonbar01a}, and actions with nondeterminism \cite{gikali97a}.


 In this survey, we characterize  planning problems using the three dimensions:  
 \begin{enumerate}
 	\item the type of action theories, 
 	\item the degree of uncertainty about the initial state, and
 	\item the availability of knowledge-producing actions. 
 \end{enumerate}
 In particular, the literature has named the following classes of planning problems.
 \emph{Classical planning} refers to planning problems with deterministic action theories and complete initial states. \emph{Conformant planning} deals with the incompleteness of the initial state and nondeterministic action theories. \emph{Conditional planning} considers knowledge producing actions and generates plans which might contain non-sequential constructs, such as \textbf{if-then} or \textbf{while loop}.

This paper presents a survey of research focused on ASP-based planning and its applications. It begins (Section \ref{sec:background}) with a brief introduction of answer set programming and action language $\mathcal{B}$,
the main representation language for planning problems. It describes different encodings of answer set planning  for problems with complete information and no sensing actions (Section \ref{asp-sec}). The paper then introduces two different approaches to planning with incomplete information (Section \ref{section:conformant}) and the description of a conditional planner, which solves planning problems in domains with sensing actions and incomplete information (Section \ref{section:sensing}). The survey explores next the problems of  planning with preferences (Section \ref{sec:prefs}), diagnosis (Section \ref{sec:diag}), planning in multi-agent environments
(Section \ref{sec:mas}), and planning and scheduling in real-world applications (Section \ref{sec:extension}). The paper concludes with a discussion about open challenges for answer set planning (Section \ref{sec:discussion}).

\section{Background}\label{sec:background}

\subsection{Answer Set Programming}

As usual, a {logic program} consists of {rules} of the form
\[
  a_1 \vee \ldots \vee a_m \leftarrow a_{m+},\dots,a_n,\naf a_{n+1},\dots,\naf a_o
\]
where each $a_i$ is
an {atom} of form $p(t_1,\dots,t_k)$
and all $t_i$ are terms,
composed of function symbols and variables.
Atoms $a_1$ to $a_m$ are often called head atom,
while $a_{m+1}$ to $a_n$
and $\naf a_{n+1}$ to $\naf a_o$
are also referred to as positive and negative body literals, respectively.
An expression is said to be {ground}, if it contains no variables.
As usual, $\naf$ denotes (default) {negation}.
A rule is called a {fact} if $m=o=1$, normal if $m=1$, and
an integrity constraint if $m=0$.
Semantically, a logic program produces a set of {stable models},
also called {answers sets},
which are distinguished models of the program determined by the stable model semantics;
see the paper by \citeN{gellif91a} for details.

To ease the use of ASP in practice,
several simplifying notations and extensions have been developed.
First of all, rules with variables are viewed as shorthands for the set of their ground instances.
Additional language constructs include
{conditional literals} and {cardinality constraints} \cite{siniso02a}.
The former are of the form $a:{b_1,\dots,b_m}$,
the latter can be written as
$s\{d_1;\dots;d_n\}t$,\footnote{More elaborate forms of aggregates can be obtained by explicitly using function (e.g.,\ \texttt{\#count}) and
	relation symbols (eg.\ \texttt{<=}).}
where $a$ and $b_i$ are possibly default-negated (regular) literals  
and each $d_j$ is a conditional literal; 
$s$ and $t$ provide optional lower and upper bounds on the number of satisfied literals in the cardinality constraint.
We refer to $b_1,\dots,b_m$ as a {condition}.
The practical value of both constructs becomes apparent when used with variables.
For instance, a conditional literal like
$a(X):b(X)$
in a rule's antecedent expands to the conjunction of all instances of $a(X)$ for which the corresponding instance of $b(X)$ holds.
Similarly,
$2\;\{a(X):b(X)\}\;4$
is true whenever at least two and at most four instances of $a(X)$ (subject to $b(X)$) are true.
Finally, objective functions minimizing the sum of a set of weighted tuples $(w_i,\boldsymbol{t}_i)$ subject to condition $c_i$ are expressed as
\(
\#minimize\{w_1@l_1,\boldsymbol{t}_1:c_1;\dots;w_n@l_n,\boldsymbol{t}_n:c_n\}
\).
Analogously, objective functions can be optimized using the $\#\textit{maximize}$ statement.
Lexicographically ordered objective functions are (optionally) distinguished via levels indicated by $l_i$.
An omitted level defaults to 0. Furthermore, 
$w_i$ is a numeral constant and $\boldsymbol{t}_i$ a sequence of arbitrary terms.
Alternatively, the above minimize statement can be expressed by weak constraints of the form
\(
\hookleftarrow c_i [w_i@l_i,\boldsymbol{t}_i]
\)
for $1\leq i\leq n$.

As an example, consider the following rule:
\[
1 \{ \mathit{move}(D,P,T) : \mathit{disk}(D), \mathit{peg}(P) \} 1 \leftarrow \mathit{ngoal}(T-1), T\leq n.
\]
This rule has a single head atom consisting of a cardinality constraint;
it comprises all instances of $\mathit{move}(D,P,T)$ where $T$ is fixed by the two body literals and $D$ and $P$ vary
over all instantiations of predicates $\mathit{disk}$ and $\mathit{peg}$, respectively.
Given 3 pegs and 4 disks, this results in 12 instances of $\mathit{move}(D,P,T)$ for each valid replacement of
$\mathit{T}$, among which exactly one must be chosen according to the above rule.

Full details of the input language of \clingo, along with various examples and its semantics, can be found
in the papers by ~\citeN{PotasscoUserGuide}. The interested reader is also referred
to the work by \citeN{cafageiakakrlemarisc19a} for the description of the ASP Core 2 Language.

 A logic program can have zero, one, or multiple answer sets.
This distinguishes answer set semantics from other semantics of logic programs
such as the well-founded semantics of \citeN{gerosc91a} or perfect models semantics of \citeN{przymusinski88b}.
Answer set semantics, together with the introduction of choice rules and constraints,
enables the development of answer set programming as proposed by \citeN{lifschitz99b}. 
Following this approach, a problem can be solved by a logic program
whose answer sets correspond one-to-one to the problem's solutions.
Choice rules are often used to generate potential solutions
and constraints are used to eliminate potential but incorrect solutions.

\subsection{Reasoning About Actions: The Action Description Language $\mathcal{B}$}
\label{sub:languageB}

We review the basics of the action description language
$\mathcal{B}$ \cite{gellif98a}.  
An action theory in $\mathcal{B}$ is defined over a set of actions \textbf{A} and a set of
fluents \textbf{F}. A fluent literal is a fluent $f \in \mathbf{F}$ or its negation $\neg f$.
A {\em fluent formula} is a Boolean formula constructed
from fluent literals.
An action theory is a  set of laws of the  form
\begin{eqnarray}
 & \caused(\varphi , f) \label{static}\\
 & \causes(a,f, \varphi) \label{dynamic}\\
 & \executable(a, \varphi) \label{exec}\\
 &  \initially(f) \label{init}
\end{eqnarray}
where $f$ and $\varphi$ are a fluent literal and a set of fluent literals, respectively, and $a$ is an
action. A law of the form \eqref{static} represents a {\em static causal law}, i.e., a
relationship between fluents. It conveys that whenever the fluent literals in $\varphi$
hold then so is $f$. A {\em dynamic causal
law} is of the form \eqref{dynamic} and  represents the (conditional) effect of $a$ while
a law of the form \eqref{exec} encodes an executability condition of $a$.
Intuitively, an executability condition of the form \eqref{exec}
states that $a$ can only be executed if $\varphi$ holds. A dynamic
law of the form \eqref{dynamic} states that $f$ is caused to be
true after the execution of $a$ in any state of the world where $\varphi$ holds.
When $\varphi = \emptyset$ in \eqref{exec}, we often omit laws of this type from the theory.
Statements of the form \eqref{init}
describe the initial state. They state that $f$ holds
in the initial state. We also often refer to  $\varphi$ as the ``\emph{precondition}''
for each particular law.

 An {\em action theory} is a pair $(D,\Gamma)$ where $\Gamma$,
called the {\em initial state}, consists of
laws of the form \eqref{init} and $D$, called {\em the
domain}, consists of laws of the form
\eqref{static}-\eqref{exec}. For convenience, we sometimes denote
the set of laws of the form \eqref{static} by $D_C$.

\begin{example}
\label{ex1}
Let us consider a modified version of the suitcase $s$
with two latches from the work by \citeN{lin95a}.
We have a suitcase with two latches $l_1$ and $l_2$.
$l_1$ is up and $l_2$ is down.
To open a latch ($l_1$ or $l_2$), we need a corresponding
key ($k_1$ or $k_2$, respectively).
When the two latches are in the up position, the
suitcase is unlocked. When one of the latches is
down, the suitcase is locked.
In this domain, we have that
\[
\mathbf{A} = \{open(l_i), close(l_i), get\_key(k_i) \mid  i \in \{1,2\}\}
\]
and
\[
\mathbf{F} = \{locked\} \cup \{up(l_i), holding(k_i)  \mid i \in \{1,2\}\}.
\]
The intuitive meaning of the actions and fluents is clear. The problem can be represented using
the  laws
\[
D^s = \left \{
\begin{array}{ll}
\causes(open(l_i), up(l_i), \emptyset) &     \\
\causes(close(l_i), \neg up(l_i), \emptyset) &     \\
\causes(get\_key(k_i), holding(k_i), \emptyset) &   \\
\executable(open(l_i), \{holding(k_i)\}) &   \\
\caused(\{\neg up(l_i)\}, locked)  &   \\
\caused(\{up(l_1), up(l_2)\}, \neg locked) &  \\
\end{array}
\right.
\]
where, in all  laws,  $i = 1,2$.
The first three laws describe the effects of the action of opening a latch, closing a latch, or getting a key.  The fourth law encodes that
we can open the latch only when we have the right key. Observe that the omission of executability laws for $close(l_i)$ or
$get\_key(k_i)$ indicates that these actions can always be executed.
The last two laws are static causal laws encoding that
the suitcase is locked
when either of the two latches is in the down position and it is unlocked when the
two latches are in the up position.

A possible initial state of this domain is given by
\[
\Gamma^s = \left \{
\begin{array}{lllll}
\initially(up(l_1)) \\
\initially(\neg up(l_2))  \\
\initially(locked)  \\
\initially(\neg holding(k_1)) \\
\initially(holding(k_2)) \\
\end{array}
\right .
\]
\hfill$\diamond$
\end{example}

 A domain given in $\mathcal{B}$  defines a transition
function from pairs of actions and states to sets of states whose
precise definition is given below. Intuitively, given an action
$a$ and a state $s$, the transition function $\Phi$ defines the
set of states $\Phi(a,s)$ that may be reached after executing the
action $a$ in state $s$. The mapping to a set of states captures the fact
that an action can potentially be non-deterministic and produce different results
(e.g., an action \emph{open} might be successful in opening a lock or not if we
account for the possibility of a broken lock). If $\Phi(a,s)$ is an empty set it means
that the execution of $a$ in $s$ is undefined. 
We now formally define $\Phi$.

 Let $D$ be a domain in $\mathcal{B}$.
A set of fluent literals is said to be {\em consistent} if it
does not contain $f$ and $\neg f$ for some fluent $f$. An
{\em interpretation} $I$ of the fluents in $D$ is a maximal
consistent set of fluent literals of $D$. A fluent
$f$ is said to be true (resp. false) in $I$ if $f \in I$ (resp.
$\neg f \in I$).
The truth value of a fluent formula in $I$ is
defined recursively over the propositional connectives in the
usual way. For example, $f \wedge g$ is true in $I$ if $f$ is
true in $I$ and $g$ is true in $I$. We say that a formula
$\varphi$ holds in $I$ (or $I$ satisfies $\varphi$), denoted by
$I \models \varphi$, if $\varphi$ is true in $I$.

 Let $u$ be a consistent set of fluent literals and $K$ a
set of static causal laws. We say that $u$ is closed under
$K$ if for every static causal law
$$\caused(\varphi,f)$$ in $K$, if
$u \models \bigwedge_{p \in \varphi} p$ then $u \models f$. By
$Cl_K(u)$ we denote the least consistent set of literals from $D$
that contains $u$ and is also closed under $K$. It is worth
noting that $Cl_K(u)$ might be undefined when it is inconsistent. For instance, if $u$
contains both $f$ and $\neg f$ for some fluent $f$, then $Cl_K(u)$
cannot contain $u$ and be consistent; another example is that if
$u = \{f,g\}$ and $K$ contains
$$\caused(\{f\}, h) \quad \quad \textnormal{ and } \quad \quad \caused(\{f, g\}, \neg h),$$
then $Cl_K(u)$ does not exist
because it has to contain both $h$ and $\neg h$, which means that
it is inconsistent.

 Formally, a {\em state} of $D$ is an interpretation of the
fluents in {\bf F} that is closed under the set of static causal
laws $D_C$ of $D$.

 An action $a$ is {\em executable} in a state $s$ if there
exists an executability proposition
$$\executable(a, \varphi)$$ in $D$ such that $s \models \bigwedge_{p \in \varphi} p$.
Clearly, if $\varphi = \emptyset$, then $a$ is executable in every state of $D$.

 The {\em direct effect of an action a} in a state $s$ is the
set
$$e(a,s) = \left \{f \mid
\causes(a, f, \varphi ) \in D, s \models  \bigwedge_{\textstyle{p \in \varphi}}
p \right \}.
$$

 For a domain $D$, the set of states $\Phi(a,s)$ that may be reached by executing
$a$ in $s$, is defined as follows.

\begin{enumerate}
\item If $a$ is executable in $s$, then
\[\Phi(a,s) = \{s'  \mid   s' \mbox{ is a state and }s' =
Cl_{D_C}(e(a,s) \cup (s
\cap s'))\}; \]

\item If $a$ is not executable in $s$, then $\Phi(a,s) = \emptyset$.

\end{enumerate}
Intuitively, the states produced by $\Phi(a,s)$ are fixpoints of an equation, obtained
by closing (with respect to all static causal laws) the set which includes the
direct effects $e(a,s)$ of action $a$ and the fluents whose value does not change as we
transition from $s$ to $s'$ through action $a$
 (i.e., $s \cap s'$).

The presence of static causal laws introduces non-determinism to action theories, i.e.,
$\Phi(a,s)$ can contain more than one element. For instance, consider the theory with the set of laws
 \[
\{\causes(a,f), \quad  \caused(\{f, \neg g\}, \neg h), \quad  \caused(\{f, \neg h\}, \neg g)\}.
 \]
It is easy to check that
\[
\Phi(a, \{\neg f, \neg g, \neg h\}) = \{\{f, g, \neg h\}, \{f, \neg g, h\}\}.
\]

 Every domain $D$ in $\mathcal{B}$ has a unique
transition function $\Phi$, and we say $\Phi$ is the transition
function of $D$. We illustrate the definition of the transition
function in the next example.

\begin{example}
\label{ex2}
{\rm
For the suitcase domain in Example \ref{ex1},
the initial state, given by the set of laws $\Gamma^s$, is defined by
\[
s_0 = \{up(l_1), \neg up(l_2),
    locked, \neg holding(k_1), holding(k_2)\}.
\]
  In state $s_0$, the three actions $open(l_2)$, $close(l_1)$,
and $close(l_2)$ are executable. $open(l_2)$ is executable since
$holding(k_2)$ is true in $s_0$ while $close(l_1)$ and
$close(l_2)$ are executable since the theory (implicitly) contains
the laws
\[
\executable(close(l_1),\{\}) \hspace{1cm} \textnormal{ and }
\hspace{1cm} \executable(close(l_2),\{\})
\]
which indicate that these two actions are always executable. The
following transitions are possible from state $s_0$:
\begin{eqnarray*}
   \{\: up(l_1), up(l_2), \neg locked,\neg holding(k_1),
    holding(k_2) \:\}
 & \in & \Phi(open(l_2), s_0).\nonumber\\
   \{\: up(l_1), \neg up(l_2), locked,\neg holding(k_1),
  holding(k_2) \:\}
 & \in & \Phi(close(l_2), s_0).\nonumber\\
   \{\: \neg up(l_1), \neg up(l_2), locked,\neg
holding(k_1),holding(k_2)
\:\}
 & \in & \Phi(close(l_1), s_0).\nonumber
\end{eqnarray*}
}
\hfill$\diamond$
\end{example}

The transition function $\Phi$ is extended to define $\widehat{\Phi}$
for reasoning about effects of action sequences in the usual way.
For a sequence of actions  $\alpha = \langle a_0,\ldots,a_{n-1} \rangle$ and a state $s$,
$\widehat{\Phi}(\alpha, s)$ is a collection of states defined as follows:

\begin{itemize}
\item $\widehat{\Phi}(\alpha, s) = \{s\}$ if $\alpha = \langle \ \rangle$;

\item $\widehat{\Phi}(\alpha, s) =  \Phi(a_0, s)$ if $n=1$;

\item $\widehat{\Phi}(\alpha, s) =  \bigcup_{s' \in \Phi(a_0, s)} \widehat{\Phi}(\alpha', s')$ if
$n > 1$, $ \Phi(a_0, s) \ne \emptyset$, and
$ \widehat{\Phi}(\alpha', s') \ne \emptyset$ for every $s' \in \Phi(a_0,s)$
where $\alpha' =  \langle a_1,\ldots,a_{n-1} \rangle$.
\end{itemize}

 A domain $D$ is \textit{consistent} if for every
action $a$ and state $s$, if $a$ is executable in $s$, then
$\Phi(a,s) \neq \emptyset$. An action theory $(D,\Gamma)$ is
consistent if $D$ is consistent and $s_0 = \{f \mid
\initially(f)\in \Gamma\}$ is a state of $D$. In what follows, we
consider only consistent action theories and refer to $s_0$ as 
the initial state of $D$.
We call a sequence of alternate states and actions $s_0a_0\ldots a_{k-1}s_k$
a \emph{trajectory} if $s_{i+1} \in \Phi(a,s_i)$ for every $i=0,\ldots,k-1$.

 A {\em planning problem} with respect to $\mathcal{B}$ is specified
by a triple $\lan D, \Gamma, \Delta \ran$ where $(D,\Gamma)$ is an
action theory in $\mathcal{B}$ and $\Delta$ is a set of fluent literals (or {\em
goal}). A sequence of actions
$\alpha = \langle a_0,\ldots,a_{n-1} \rangle$ is then called an {\em optimistic plan for
$\Delta$} if  there exists some $s \in \widehat{\Phi}(\alpha, s_0)$
such that $s \models \Delta$ where $s_0$ is the initial state of $D$. Note that we use the term `optimistic plan' to refer to $\alpha$,
as used by \citeN{eifalepfpo03b}, instead of `plan' because the non-determinicity of $D$
does not guarantee that the goal is achieved in every state reachable after the execution of $\alpha$.
However, if $D$ is deterministic, i.e., $|\Phi(a,s)| \le 1$ for every pair
$(a,s)$ of actions and states,
then the two notions of `optimistic plan' and `plan' coincide.



\section{Planning with Complete Information}
\label{asp-sec}

Given a planning problem $\mathcal{P} = \lan D,\Gamma, \Delta \ran$,
answer set planning solves it by translating it into a logic program $\Pi(\mathcal{P}, n)$
whose answer sets correspond one-to-one to optimistic plans of length $\le n$ of $\mathcal{P}$.
Intuitively, each answer set corresponds to a trajectory $s_0a_0\ldots a_{n-1}s_n$
such that  $s_n \models \Delta$. As such, the choices that need to be made at each step $k$
in program $\Pi(\mathcal{P}, n)$ are either the action $a_k$ or the state $s_k$.
This leads to different encodings, referred to as \emph{action-based} and \emph{state-based},
which emphasize the object of the selection process.

Over the years, several types of encodings have been developed. We present below two of the
most popular mappings of planning problems to logic programs.
The first encoding, presented in Subsection \ref{direct}, views the problem $\mathcal{P}$ as a set of laws
in the language $\mathcal{B}$ (Subsection~\ref{sub:languageB})
while the second encoding, illustrated in Subsection \ref{facts}, views the problem  as a set of facts. In both encodings, the program
$\Pi(\mathcal{P}, n)$ contains the atom\footnote{
   Throughout the paper, we will use the syntax implemented in the \clingo\ system.
} $time(0..n)$ to represent  the set of facts $\{time(0), \ldots, time(n)\}$.


\subsection{A Direct Encoding}\label{direct}

The rules of $\Pi(\mathcal{P}, n)$ in this encoding are described by \citeN{sobanamc03a}.
The main predicates in the program are:

\begin{enumerate}
\item $holds(F,T)$ -- the fluent literal $F$ is true at  time step $T$;
\item $occ(A,T)$ -- the action $A$ occurs at  time step $T$; and
\item $possible(A,T)$ -- the action $A$ is executable at  time step $T$.
\end{enumerate}
The program contains two sets of rules. The first set of rules is domain dependent.
The rules in the  second set  are generic and common to  all problems.
For a set of literals $\varphi$, we use
$holds(\varphi, T)$ to denote the set $\{holds(L, T) \mid L \in \varphi\}$.

\subsubsection{Domain-Dependent Rules}\label{subdep}

For each planning problem
$\mathcal{P} = \lan D,\Gamma, \Delta \ran$,  program $\Pi(\mathcal{P}, n)$ contains the following rules:

\begin{enumerate}
\item For each element $\initially(f)$ of   form \eqref{init} in $\Gamma$,  the fact
\begin{equation} \label{ir_init}
holds(f, 0) \leftarrow
\end{equation}
stating  that the fluent literal $f$ holds at  time step $0$.

\item For each executability condition $\executable(a, \varphi)$ of   form \eqref{exec} in $D$, the rule
\begin{equation} \label{ir_pos}
possible(a, T) \leftarrow time(T), holds(\varphi,T)
\end{equation}
stating that it is possible to execute the action $a$ at
 time step $T$ if $\varphi$ holds at time step $T$.

\item For each dynamic causal law $\causes(a,f, \varphi)$ of   form \eqref{dynamic}  in $D$, the rule
\begin{equation} \label{ir_dyn}
\begin{array}{lll}
holds(f, T+1) & \leftarrow & time(T), occ(a, T), holds(\varphi, T)
\end{array}
\end{equation}
stating  that if
$a$ occurs at  time step $T$
then the fluent literal $f$ becomes true at $T+1$ if the conditions in
$\varphi$ hold.

\item For each static causal law $\caused(\varphi , f)$ of   form \eqref{static}  
in $D$,  the  rule\footnote{
      If $f = \mathit{false}$ then the head of the rule is empty.
}
\begin{equation} \label{ir_sta}
holds(f, T) \leftarrow time(T), holds(\varphi,T)
\end{equation}
which represents a straightforward translation of the static causal
law into a logic programming rule.

\item To guarantee that an action is executed only when it is executable, the constraint
\begin{equation} \label{ir_constraint}
\leftarrow time(T), occ(A, T), \naf possible(A,T)
\end{equation}

\end{enumerate}

We demonstrate the above translation by listing the set of domain dependent rules for the domain from Example \ref{ex1}.
\begin{example}
\label{ex4}
Besides the set of facts encoding actions and fluents and the initial state
(for each $a \in \mathbf{A}$, $f \in  \mathbf{F}$ and
$\initially(l)\in \Gamma^s$),
$$
action(a) \leftarrow   \hspace{2cm}  fluent(f) \leftarrow   \hspace{2cm} holds(l,0) \leftarrow
$$
the encoding of the suitcase domain in Example \ref{ex1} contains
the following rules
for $i=1,2$:
\[
\begin{array}{lll}
holds(up(l_i), T+1)                     & \leftarrow & time(T), occ(open(l_i), T) \\
holds(\neg up(l_i), T+1)                & \leftarrow & time(T), occ(close(l_i), T) \\
holds(holding(k_i), T+1)        & \leftarrow & time(T), occ(get\_key(k_i), T) \\
possible(open(l_i), T)                  & \leftarrow & time(T), holds(holding(k_i), T) \\
holds(locked, T)                        & \leftarrow & time(T), holds(\neg up(l_i), T) \\
holds(\neg locked, T)                   & \leftarrow & time(T), holds(up(l_1), T), holds(up(l_2), T) \\
possible(close(l_i), T)                 & \leftarrow & time(T) \\
possible(get\_key(k_i), T)              & \leftarrow & time(T) \\
\end{array}
\]
Each of the first six rules corresponds to a law in $D^s$. The last two rules are added
because there is no restriction on the executability condition of $close(l_i)$ or $get\_key(k_i)$.
\hfill $\Diamond$
\end{example}

\subsubsection{Domain Independent Rules} \label{subind}

The set of domain independent rules of $\Pi(\mathcal{P}, n)$  consists of rules
for generating action occurrences and encoding the frame axiom.

\begin{enumerate}

\item \emph{Action generation rule:} To create plans, $\Pi(\mathcal{P}, n)$
must contain rules that generate action occurrences of the form $occ(a,t)$. This is encoded by the rule
\begin{eqnarray}
1 \{ occ(A,T) : action(A) \} 1 \leftarrow time(T), T < n \label{occ}
\end{eqnarray}
stating that exactly one action must occur at each time step.
It makes use of  the  cardinality atom $1 \{occ(A,T) : action(A) \} 1$
which is true for a time step $T$ iff exactly one atom in the set $\{occ(A,T) : action(A) \}$ is true.
The former atom can be replaced by
  $l \{occ(A,T) : action(A) \} u$ where $0 \le l \le u$ to allow for different types of plans, e.g., for parallel
  plans, $l>1$.

\item \emph{Inertia rule:}  The frame axiom, which states that a property   continues to hold unless it is changed,
is encoded as follows:
\begin{eqnarray}
holds(F, T{+}1)         & \leftarrow & time(T), fluent(F), holds(F, T), \naf holds(\neg F, T{+}1) \quad \quad \label{inertial_1} \\
holds(\neg F, T{+}1)    & \leftarrow & time(T), fluent(F), holds(\neg F, T), \naf holds(F, T{+}1) \label{inertial_2}
\end{eqnarray}

\item \emph{Consistency constraint}: To ensure that states encoded by
answer sets are consistent,  $\Pi(\mathcal{P}, n)$ contains the following constraint:
\begin{eqnarray}
   & \leftarrow & fluent(F),
holds(F, T), holds(\neg F, T)  \label{constraint}
\end{eqnarray}
Observe that this constraint is needed because $holds(f,t)$ and $holds(\neg f, t)$ are
``consistent'' for answer set solvers. It would not be needed if $holds(\neg f, t)$ is encoded
as $\neg holds(f, t)$.

\end{enumerate}

\subsubsection{Goal Representation} \label{subgoal}

The goal $\Delta$ is encoded by rules defining the predicate $goal$,
which is true whenever $\Delta$ is true, and a rule that enforces
that $\Delta$ must be true at   time step $n$.
For example, if $\Delta$ is  a conjunction of literals  $p_1 \wedge \ldots \wedge p_k$, then the rules
\begin{eqnarray}
goal  & \leftarrow  & holds(p_1, n), \ldots, holds(p_k, n) \label{goal} \\
        & \leftarrow & \naf goal \label{enforce_goal}
\end{eqnarray}
encode $\Delta$ and enforce that $\Delta$ must be  true at time step $n$.

\subsubsection{Correctness of the Encoding}
\label{correctness-pi}

Let $\mathcal{P} = (D,\Gamma,\Delta)$ and $\Pi(\mathcal{P},n)$ be the logic
program consisting of
\begin{itemize}
\item the set of facts encoding fluents and literals in $D$;
\item the set of domain-dependent rules encoding $D$ and $\Gamma$
(rules \eqref{ir_init}--\eqref{ir_constraint})
in which the domain of $T$ is $\{0,\ldots,n\}$;
\item the set of domain-independent rules
(rules \eqref{occ}--\eqref{inertial_2})
in which the domain of $T$ is $\{0,\ldots,n\}$; and
\item the rules \eqref{constraint}--\eqref{enforce_goal}.
\end{itemize}
The following result shows the equivalence between optimistic plans
achieving $\Delta$ and answer sets of $\Pi(\mathcal{P},n)$. To formalize the
theorem, we introduce some additional notation.
For an answer set
$M$ of $\Pi(\mathcal{P},n)$, we define
\[
s_i(M) = \{f \  \mid f \textnormal{ is a fluent literal and }
holds(f,i) \in M\}.
\]
\begin{theorem}
\label{th1}
For a planning problem  $\lan D,\Gamma,\Delta \ran$ with a consistent
action theory $(D,\Gamma)$, $s_0a_0\ldots a_{n-1}s_n$ is a trajectory achieving $\Delta$ iff
there exists an answer set $M$ of  $\Pi(\mathcal{P},n)$ such that
\begin{enumerate}
\item $occ(a_i,i) \in M$ for  $i \in \{0,\ldots,n-1\}$ and
\item
$s_i = s_i(M)$ for $i \in \{0,\ldots,n\}$.
\end{enumerate}
\end{theorem}
\begin{remark}
\label{remark:classical_planning}
\begin{enumerate}
\item The proof of Theorem~\ref{th1} relies on the following observations:
$M$ is an answer set of $\Pi(\mathcal{P},n)$ iff
\begin{itemize}
\item for every $i$ such that  $0 \le i < n$, there exists some $a_i \in \mathbf{A}$ such that

$occ(a_i, i) \in M$ and $a_i$ is executable in $s_i(M)$;\item $s_0(M)$ is the initial state of the action theory $(D,\Gamma)$ and is consistent. Furthermore,
for every $i$ such that $0 \le i < n$,
$s_{i+1}(M) \in \Phi(a_i, s_i(M))$; and
\item $\Delta$ is true in $s_n(M)$.
\end{itemize}
The theorem is similar to the correspondence between histories and answer sets explored by
\citeN{liftur99a} and by \citeN{sobanamc03a}.

\item If $(D,\Gamma)$ is deterministic then Theorem~\ref{th1} can be simplified to ``\emph{
$a_0,a_1,\ldots,a_{n-1}$ is a plan achieving $\Delta$ iff
there exists an answer set $M$ of  $\Pi(\mathcal{P},n)$ such that
$occ(a_i,i) \in M$ for  $i \in \{0,\ldots,n-1\}$.}''

\item\label{item-direct-encoding}
A different variant of this encoding, which uses $f(\vec{x}, t)$ and $\neg f(\vec{x}, t)$ instead of $holds(f(\vec{x}), t)$
and $holds(\neg f(\vec{x}), t)$, respectively, can be found in several papers related to answer set planning, e.g.,
in the papers by \citeN{lifschitz99b} and \citeyear{lifschitz02a}.

\item The action language $\mathcal{B}$ could be extended with various features such as default fluents, effects
of action sequences, etc. as discussed by \citeN{gellif98a}. These features can be easily included in the encoding
of  $\Pi(\mathcal{P},n)$. On the other hand, such features are rarely considered in action domains used by the
planning community. For this reason, we do not consider such features in this survey.

\item Readers familiar with current answer set solvers such as \clingo\ or \dlv\ could be wondering why
$holds(\neg f, t)$ is used instead of a perhaps more intuitive $\neg holds(f,t)$. Indeed, the former can be replaced by the latter.
The use of $holds(\neg f, t)$ is influenced by early Prolog programs written for reasoning about actions and changes
by Michael Gelfond. A Prolog program that translates a planning problem to its ASP encoding
can be found at {\footnotesize \url{https://www.cs.nmsu.edu/~tson/ASPlan/Knowledge/translate.pl}}.

\item Different approaches to integrate various types of knowledge to answer set planning can be found in the work 
by \citeN{dikuna05a} and \citeN{sobanamc03a}.

\end{enumerate}
\end{remark}

\subsection{Meta Encoding}\label{facts}

The meta encoding presented in this section encodes a planning problem $\mathcal{P} = \langle D, \Gamma, \Delta \rangle$
as a set of facts, in addition to  a set of domain-independent rules for reasoning about effects of actions. To distinguish
this encoding from the previous one, we denote this encoding with $\Pi^m(\mathcal{P},n)$.
In this encoding, a set $\varphi$ is represented using the atom $set(s_\varphi)$, where $s_\varphi$ is a new atom associated to $\varphi$, and
a  set of atoms of the form $\{member(s_\varphi, p) \mid p \in \varphi \}$. The laws in $D$ are 
represented by the set of facts
\begin{eqnarray}
caused(f, s_{\mathit{sf}}).    & set(s_{\mathit{sf}}).  & s_{\mathit{sf}}  \textnormal{ is the identifier for  } \varphi  \textnormal{ in  } \caused(f, \varphi) \textnormal{ in } D \\
causes(a,  f, s_{\mathit{df}}).& set(s_{\mathit{df}}). & s_{\mathit{df}} \textnormal{ is the identifier for  } \varphi  \textnormal{ in  } \causes(a, f, \varphi) \textnormal{ in } D \\
executable(a,  s_a). & set(s_a). & s_{a} \textnormal{ is the identifier for  } \varphi  \textnormal{ in  } \executable(a, \varphi) \textnormal{ in } D
\end{eqnarray}
and the set of facts encoding $s_{\mathit{sf}}$, $s_{\mathit{df}}$, and $s_a$.

In addition to the action generation rule \eqref{occ}, the inertial rules \eqref{inertial_1}--\eqref{inertial_2}, the goal representation
rules \eqref{goal}-\eqref{enforce_goal}, and the constraint stating that actions can occur only when they
are executable \eqref{ir_constraint}, the program $\Pi^m(\mathcal{P},n)$ contains the following rules for reasoning about effects of actions:
%
\begin{eqnarray}
holds(S, T)  & \leftarrow       & time(T), set(S),   \label{set_hold} \\
                   &                    & holds(F,T): fluent(F), member(S,F); \nonumber \\
                    &                   & holds(\neg F, T): fluent(F), member(S, \neg F).   \nonumber \\
holds(L, T+1)  & \leftarrow & time(T), causes(A, F, S), occ(A, T), holds(S, T)    \label{m_dyna} \\
holds(L, T)      & \leftarrow & time(T), caused(L, S), holds(S, T)    \label{m_stat} \\
possible(A, T) & \leftarrow & time(T), executable(A, S),  holds(S, T)    \label{m_exec}
\end{eqnarray}
Rule \eqref{set_hold} defines the truth of a set of fluents $S$  at  time step $T$, by declaring that $S$ holds at $T$ if
all of its members are true at $T$.
The intuition behind the rules \eqref{m_dyna}--\eqref{m_exec} is clear. Similarly to Theorem~\ref{th1},
answer sets of $\Pi^m(\mathcal{P},n)$ correspond one-to-one to possible solutions (optimistic plans) of $\mathcal{P}$.

\begin{remark}
A similar encoding to $\Pi^m(\mathcal{P},n)$ is used in the system \plasp\ version~3 by \citeN{digelurosc18a}. A translation
of planning problems from PDDL format to ASP facts can be found at  {\footnotesize \url{https://github.com/potassco/plasp}}.
\end{remark}

\subsection{Adding Heuristics: Going for Performance}
\label{subsection:heuristic}

By using ASP systems for solving planning problems, we employ general-purpose systems rather than genuine planning systems.
In particular, the distinction between action and fluent variables or fluent variables of successive states
completely eludes the ASP system.
Pioneering work in this direction was done by \citeN{rintanen12a}, where the
implementation of SAT solvers was modified in order to boost performance of SAT planning.
Inspired by this research direction,
\citeN{gekaotroscwa13a} developed a language extension for ASP systems that allows users to
declare heuristic modifiers
that take effect in the underlying ASP system \clingo.

More precisely,
a heuristic directive is of form
\[
\#\mathit{heuristic}\ a:b_1,\dots,b_m.\,[w,m]
\]
where $a$ is an atom and $b_1,\dots,b_m$ is a conjunction of literals;
$w$ is a numeral term and $m$ a heuristic modifier,
indicating how the solver's heuristic treatment of $a$ should be changed whenever $b_1,\dots,b_m$ holds.
\Clingo\ distinguishes four primitive heuristic modifiers:
\begin{description}
\item [\textit{init}] for initializing the heuristic value of $a$ with $w$,
\item [\textit{factor}] for amplifying the heuristic value of $a$ by factor $w$,
\item [\textit{level}] for ranking all atoms; the rank of $a$ is $w$,
\item [\textit{sign}] for attributing the sign of $w$ as truth value to $a$.
\end{description}
For instance, whenever $a$ is chosen by the solver,
the heuristic modifier \textit{sign} enforces that it becomes either true or false
depending on whether $w$ is positive or negative, respectively.
The other three modifiers act on the atoms' heuristic values assigned by the ASP solver's heuristic function.%
\footnote{See the paper by ~\citeN{gekaotroscwa13a} and the user guide by \citeN{PotasscoUserGuide} for a comprehensive introduction to heuristic modifiers in \clingo.}
Moreover,
for convenience,
\clingo\ offers the heuristic modifiers \textit{true} and \textit{false} that
combine \textit{level} and \textit{sign} statement.

With them,
we can directly describe the heuristic restriction used in the work by \citeN{rintanen11a} to simulate planning
by iterated deepening $A^*$ \cite{korf85a} through limiting choices to action variables,
assigning those for time \texttt{T} before those for time \texttt{T+1}, and always assigning truth
value \texttt{true} (where \texttt{n} is a constant indicating the planning horizon):
\[
\#\mathit{heuristic}\ \mathit{occ}(A,T) : \mathit{action}(A), \mathit{time}(T).\, [n-T,\mathit{true}]
\]

Inspired by, and yet different from the work by \citeN{rintanen12a},
\citeN{gekaotroscwa13a} devise a dynamic heuristic that aims at propagating fluents' truth values backwards in time.
Attributing levels via \texttt{n-T+1} aims at proceeding depth-first from the goal fluents.
\begin{align*}
\#\mathit{heuristic}\ \mathit{holds}(F,T-1) &: \mathit{holds}(F,T).\, [n-T+1,\mathit{true}]\\
\#\mathit{heuristic}\ \mathit{holds}(F,T-1) &: fluent(F), time(T), not \mathit{holds}(F,T).\, [n-T+1,\mathit{false}]
\end{align*}
In an experimental evaluation conducted by \citeN{gekaotroscwa13a},
this heuristic led to a speed-up of up to two orders of magnitude on satisfiable planning problems.

\subsection{Context: Classical Planning}

Classical planning has been an intensive research area for many years.
The famous Shakey robot\footnote{\url{http://www.ai.sri.com/shakey/}} used a planner for path planning.
This project also led to the introduction of the
\emph{Stanford Research Institute Problem Solver (STRIPS)} language \cite{fiknil71a}, the first representation language for planning domain description. This language has since evolved into
the \emph{Planning Domain Definition Language (PDDL)} \cite{ghhoknmcravewewi98a}, a major planning domain description language.
We noted that PDDL with state constraints is as expressive as $\mathcal{B}$.
It is worth noticing that state constraints are often not considered by the planning community even though the benefit of dealing directly with state constraints is known \cite{thhone03a}. Furthermore, it is often assumed that state constraints in PDDL are stratified, e.g., the dependency graph among fluent literals\footnote{
   The dependency graph is a directed graph whose nodes are fluent literals and whose set of edges contains $(l, l')$ if $\caused(\varphi, l)$ is a static causal law and $l' \in \varphi$.
} should be cycle free.

Several planning algorithms have been developed and implemented such as forward or backward search over the state space (see, a survey by \citeN{hetadr90a}) and search in the plans space (a.k.a. partial order planning, see, e.g. a survey by \citeN{weld94a}).
Such research also led to the development of domain-dependent planners which utilize domain knowledge to improve their scalability and efficiency
(e.g., hierarchical planning systems \cite{sacerdoti74a}).
Researchers realized early on  that systematic state space search would  not yield planning systems
that are sufficiently scalable and efficient for practical applications.
A significant milestone in the development of domain-independent planners is the invention of {\sc GraphPlan} by \citeN{blufur97a}.
The basic data structure of {\sc GraphPlan}, the \emph{planning graph}, is
 an important resource for the development of planning heuristics \cite{kapala97a}.
It plays a key role in the success of heuristic planners such as
{\sc FF} \cite{hofneb01a} and {\sc HSP} \cite{bongef01a} which dominate several International Planning Competitions \cite{lokasebogekobrhorianwesmfo00a,bacchus01a,ipc5}.
This success is followed by several other systems \cite{helmert06a,richel09a,helmertrs+11}, whose impressive performance can be attributed to
advances in the representation language for planning (e.g., the language SAS$^+$ that supports a compact representation of states \cite{bacneb95a}) and their underlying heuristics constructed via reachability analysis and techniques such as landmarks recognition, abstraction, operator ordering, and decomposition  \cite{bonhel10a,zhugiv04a,heldom09a,helgef08a,helmat08a,hopose04a,hoffmann05a,porohecarosa20a,richel09a,roghel10a,vidgef06a}. All of these planning systems implement a heuristic search algorithm. Therefore, their scalability and efficiency are heavily dependent on the implemented heuristic, i.e., how discriminant  is the heuristic and how efficient can it be computed. In most systems, completeness and efficiency have to be traded off. In some planner, an automatic mechanism for returning to systematic search is established whenever the heuristic deems not useful (e.g., the system {\sc FF}).  
 
The idea of using automated theorem solvers in planning can be traced back to the work by \citeN{green69a} who demonstrated that automated reasoning systems can be used for planning. 
A significant step in this direction is proposed by
\citeN{kausel92b} who introduced satisfiability planning and showed that with an improved satisfiability solver, SAT-based planning can be competitive with search based planners \cite{kamcse96a}. 
This approach was later advanced by several other researchers and results in many SAT-based or logic programming-based planning systems \cite{chhuxizh09a,dineko97a,riheni06a,rogrphsa09a,rintanen12a} that are often  competitive or more efficient comparing to search-based planners. Constraint satisfaction techniques have also been employed in planning \cite{kauwal99a,dokam03a,siddim10a,dofopo09a}. 
As we have mentioned in the introduction, the success of SAT-base planning is likely the source of inspiration for the use of logic programming with answer sets semantics for planning. Indeed, there are several similarities between a SAT-based encoding of a planning program proposed by \citeN{kausel92b} and its ASP-encoding presented in this section. They share the following features:  
\begin{itemize} 
\item the use of time steps in representing the planning horizon: SAT-based encoding prefers to use $f(\vec{x},t)$ and $\neg f(\vec{x},t)$ instead of 
$holds(f(\vec{x}),t)$ and $holds(\neg f(\vec{x}),t)$;  

\item the encoding of actions' executability and the effects of actions;  

\item the encoding of the frame axioms; and   

\item the encoding for action generation. 
\end{itemize} 
In this sense, one can say that SAT-planning and answer set planning are cousins to each other. 
Both relish the use of knowledge representation techniques and the development of logical solvers in planning.  
The key difference between them lies in the underlying representation language and solver.


Green's idea has also been investigated in event calculus planning.
The main reasoning system behind this approach is the event calculus, which is introduced by \citeN{kowser86a} for reasoning about narratives and database updates.
An action theory (or a planning problem) can be described by an event calculus program that is similar to the program described in Section~\ref{direct}.
In particular, this program consists of rules encoding the initial state, effects of actions, and solution to the frame axiom.
Earlier development of event calculus does not consider static causal laws.
This issue is addressed by \citeN{shanahan99a}.
\citeN{eshghi88a} introduced a variant of the event calculus, called  \emph{EVP} (an event calculus for planning), and combined it with abductive reasoning to create ABPLAN.
We believe that this is the first planning system that integrates event calculus and abduction.
\citeN{demibr92a} developed SLDNFA, a procedure for temporal reasoning with abductive event calculus, and showed how this procedure can be used for planning.
The authors of SLDNFA continued this line of research and developed CHICA \cite{mibrde95a}.
The underlying algorithm of this system is a specialized version of the abductive reasoning procedure for event calculus.
An interesting feature of this system is that it allows for the user to specify the search strategy and heuristics at the domain level, allowing for domain dependent information to be exploited in the search for a solution.
Other proof procedures for event calculus planning can be founded in the work by \citeN{enmasateto04a}, \citeN{mueller06a}, and \citeN{shanahan00a} and \citeyear{shanahan97a}.
It is worth noting that a major discussion in these work is the condition for the soundness and completeness of the proof procedures, i.e., the planning systems. 
To the best of our knowledge, most of the event calculus based  planning systems are implemented on a Prolog system and no experimental evaluation against other planning systems has been conducted.





\section{Conformant Planning}
\label{section:conformant}

The previous section assumes that the initial state $\Gamma$ in the planning problem $\mathcal{P} = \langle D,\Gamma, \Delta \rangle$ is complete, i.e., the truth value of each property of the world is known. In practice, this is not always a realistic assumption.

\begin{example}
[Bomb-In-The-Toilet Example]
\label{example:bomb}
There may be
a bomb in a package. The process of dunking the package
into a toilet will disarm the bomb. This action can be executed
only if the toilet is not clogged. Flushing
the toilet will unclog it. This domain can be described by the
following domain:
\begin{itemize}
\item Fluents: $armed, clogged$
\item Actions: $dunk, flush$
\item Domain description:
\[
D_b = \left \{
\begin{array}{lll}
\causes(dunk ,  \neg armed, \{ armed\}) \\
\causes(flush,  \neg clogged, \{\}) \\
\executable(dunk, \{\neg clogged\})
\end{array}
\right .
\]
\end{itemize}
Suppose that our goal is to disarm the bomb. However, we are not sure whether the toilet is clogged.
In other words, the planning problem that we need to solve
is $\mathcal{P}_{bomb} = \langle D_{bomb}, \emptyset, \neg armed\rangle$.
\hfill $\Diamond$
\end{example}

The problem $\mathcal{P}_{bomb}$ is an example of a planning problem with incomplete information.
It is easy to see that $\alpha= \langle dunk \rangle$ is not a good solution for $\mathcal{P}_{bomb}$
since $\alpha$ is not executable when the toilet is clogged.
On the other hand, $\beta= \langle flush, dunk \rangle$ is executable and achieves the goal in \emph{every possible} initial state
of the problem. \citeN{eifalepfpo03b} refer to $\beta$ as a \emph{secure} plan---a solution---for the conformant planning
problem $\mathcal{P}_{bomb}$.

Let $D$ be an action theory and $\delta$ a set of fluent literals of $D$. We say that $\delta$
is a partial state of $D$ if there exists some state $s$ such that $\delta \subseteq s$.
$comp(\delta)$, called the \emph{completion} of $\delta$, denotes the set of all states $s$ such that
$\delta \subseteq s$.
A literal $\ell$ \emph{possibly holds} in $\delta$ if $\delta \not\models \overline{\ell}$ where $\overline{\ell}$ denotes the complement of $\ell$.
A set of literals $\lambda$ possibly holds in $\delta$ if every element of $\lambda$ possibly holds in $\delta$. 
In the following, we often use superscripts and subscripts to differentiate partial states from states.

A {\em conformant planning problem} $\mathcal{P}$ is a tuple $\langle {D},\delta^0,\Delta \rangle$ where $D$
is an action theory and $\delta^0$ is a partial state of ${D}$. 

An action sequence  $\alpha = \langle a_0,\dots,a_{n-1}\rangle$
is a {\em solution}  (or \emph{conformant/secure} plan) of $\mathcal{P}$ if for every state
$s_0 \in comp(\delta^0)$,  $\widehat{\Phi}(\alpha, s_0) \ne \emptyset$
and $\Delta$ is true in every state belonging to $\widehat{\Phi}(\alpha, s_0) \ne \emptyset$.

Observe that  conformant planning belongs to a higher complexity class than classical planning (see, e.g., the work by \citeN{bakrtr00a}, \citeN{eifalepfpo00a}, \citeN{hasjon99a}, or \citeN{turner02a}).
Even for action theories without static causal laws, checking whether  a conformant problem has a polynomially bounded solution is $\Sigma^2_P$-complete. Therefore, simply modifying the rules encoding the initial state \eqref{ir_init} of $\Pi(\mathcal{P},n)$ (e.g., by adding rules to complete the initial state) is insufficient. Different approaches have been proposed for conformant planning, each addressing the incomplete information in the initial state in a different way. In this section, we discuss two ASP-based approaches proposed by  \citeN{eifalepfpo03b}, \citeN{sotugemo05a}, and \citeN{tusogemo11a}.

\subsection{Conformant Planning With A Security Check Using Logic Program}
\label{subsection:dlvk}

\citeN{eifalepfpo03b} introduced the system \dlvk{} for planning with incomplete information.
The system employs a representation language that is richer than the language $\mathcal{B}$,
 since it considers additional features such as defaults
 and effects after a sequence of actions. For simplicity of the presentation, we present the approach
used in \dlvk{} for conformant planning problems specified in $\mathcal{B}$. We note that the original
\dlvk{} employs the direct encoding of planning problems as described in Remark~\ref{remark:classical_planning}, Item~\ref{item-direct-encoding}.
We adapt it to the encoding used in the previous section.

\dlvk{} generates a conformant plan for a problem $\mathcal{P}$ in two steps.
The first step consists of generating an optimistic plan; the second step is
the verification that such plan is a secure plan,
since an optimistic plan is not necessarily a secure plan. \dlvk{} implements Algorithm~\ref{dlvk-algorithm}.

\begin{algorithm2e}[htbp]
\DontPrintSemicolon
\KwIn{Conformant planning problem $\mathcal{P} = \langle {D},\delta^0,\Delta \rangle $} 
\KwOut{A secure plan $\alpha$ for $\mathcal{P} $ }
{
\While{there exists an optimistic plan $\alpha$ for $\mathcal{P}$}
{
    \If{ $\alpha$ is a secure plan} {
         \Return $\alpha$
    }
}
 {\Return no-plan}
}
\caption{\dlvk{} Algorithm}
\label{dlvk-algorithm}
\end{algorithm2e}

The two tasks in Lines 1 and 2 in Algorithm~\ref{dlvk-algorithm} are implemented using different ASP programs.
The generation step (Line 1) can be done using the program $\Pi_{c_{dlv}}(\mathcal{P},n)$ which consists of   
the program $\Pi(\mathcal{P},n)$ together with the rules that specify the values
of unknown fluents in the initial state:
\begin{equation} \label{init-incomplete}
holds(f, 0) \vee holds(\neg f, 0) \leftarrow \quad\quad (f \textnormal{ is a fluent and } \{f, \neg f\} \cap \delta^0 = \emptyset)
\end{equation}
It is easy to see that any answer set of $\Pi_{c_{dlv}}(\mathcal{P},n)$ contains an optimistic plan (Theorem~\ref{th1}).
Assume that $\alpha = \langle a_0,\ldots,a_{n-1} \rangle$ is the sequence of actions generated by  
program $\Pi_{c_{dlv}}(\mathcal{P},n)$,
\dlvk{} takes $\alpha$ and the action theory $(D, \delta^0)$ and creates a program that checks whether or not
$\alpha$ is a secure plan. If it is, then \dlvk{} returns $\alpha$. Otherwise, it continues
computing an optimistic plan and verifying that the plan is secure, until there are no
additional  optimistic plans available, which indicates that the problem has no solution. We next discuss the main
idea in the second step of \dlvk{} (Line 2).

Intuitively, if $\alpha$ is a secure plan, then its execution in every possible initial state results in a final state
in which the goal is satisfied. In other words, $\alpha$ is not a secure plan if there exists an initial state in which  the execution of $\alpha$ is not possible. This can happen in the following situations:
\begin{itemize}

\item the goal is not satisfied after the execution of $\alpha$;

\item some action $a_i$ in $\alpha$ is not executable, i.e., $possible(a_i, i)$ is not true; or

\item some constraints are violated.
\end{itemize}
Let $Check(\mathcal{P}, \alpha, n)$ be the program obtained from $\Pi_{c_{dlv}}(\mathcal{P},n)$ by
introducing a new atom, $notex$, which denotes that $\alpha$ is not secure and
\begin{itemize}
\item replacing \eqref{ir_constraint} with the rule
\[
notex \leftarrow time(T), occ(A, T), \naf possible(A,T)
\]

\item replacing \eqref{occ} with the set of action occurrences
\[
\{occ(a_i,i) \mid i=0,\ldots,n-1\}
\]

\item replacing \eqref{constraint} with
\[
\begin{array}{rcl}
	notex        & \leftarrow & fluent(F), T > 0, holds(F, T), holds(\neg F, T) \\
 			& \leftarrow &   fluent(F), holds(F, 0), holds(\neg F, 0)
\end{array}
\]

\item replacing \eqref{ir_sta}, for each constraint $\caused(\varphi, \mathit{false})$, with the rule
\[
notex  \leftarrow time(T), T > 0, holds(\varphi,T)
\]

\item replacing \eqref{enforce_goal} with
\[
\leftarrow goal, \naf notex
\]
\end{itemize}
If $Check(\mathcal{P}, \alpha, n)$ has an answer set $M$ then either the goal is not
satisfied or $notex \in M$. Observe that the rules replacing \eqref{constraint} and \eqref{ir_sta} guarantee that
$\{f \mid holds(f, 0) \in M\} \cup \{\neg f \mid holds(\neg f, 0) \in M\}$ is a state of the action domain
in $\mathcal{P}$.
If the goal is not satisfied and $notex\not\in M$,  then we have found
an initial state from which the execution of $\alpha$ does not achieve the goal.
Otherwise, $notex\in M$ implies that
\begin{itemize}
	\item  an action $a_i$ is not executable
	in the state at time step $i$ or
	\item  there is some step $j > 0$ such that the state at time step $j$ is inconsistent
	or violates some static causal laws.
\end{itemize} 
In either case, this means that there exists some initial state
in which the execution of $\alpha$ fails, i.e., $\alpha$ is not a secure plan.  On the other hand, if
$Check(\mathcal{P}, \alpha, n)$ has no answer sets, then  there are no possible initial
states such that the execution of $\alpha$ fails, i.e., $\alpha$ is a secure plan.

\subsection{Approximation-Based Conformant Planning}

Approximation-based conformant planning deals with the complexity of conformant planning
by proposing a deterministic approximation of the transition function $\Phi$, denoted by $\Phi^a$,
which maps pairs of actions and partial states such that for every $\delta$, $\delta'$,
and $s$ such that $\Phi^a(a, \delta)  = \delta'$ and $ s \in comp(\delta)$, it holds that
\begin{itemize}
\item $a$ is executable in $s$; and
\item  $\delta' \subseteq s'$ for every $s' \in \Phi(a, s)$.
\end{itemize}
Intuitively, the above conditions require $\Phi^a$ to be {\em sound} with respect to $\Phi$.
We say that $\Phi^a$ is \emph{sound approximation} of $\Phi$ if the above conditions are satisfied.

The function
$\Phi^a$ is extended to define $\widehat{\Phi^a}$ in the similar fashion as $\widehat{\Phi}$:
for an action sequence $\alpha = \langle a_0, \ldots, a_{n-1} \rangle$,
\begin{itemize} 
\item $\widehat{\Phi^a}(\alpha, \delta) = \delta$ if $\alpha = \langle \ \rangle$;
\item $\widehat{\Phi^a}(\alpha, \delta) =  \Phi^a(a_0, \delta)$ for $n=1$; and 
\item 
$\widehat{\Phi^a}(\alpha, \delta) = \widehat{\Phi^a}(\alpha', \Phi^a(a_0, \delta))$
where $\alpha' = \langle a_1, \ldots, a_{n-1} \rangle$, if
$\widehat{\Phi^a}(\beta, \delta)$ is defined for every prefix $\beta$ of $\alpha$.
\end{itemize} 

Given a sound approximation $\Phi^a$, we have that if $\widehat{\Phi^a}(\alpha, \delta) = \delta'$
then for every $ s \in comp(\delta)$, $\widehat{\Phi}(\alpha,s) \ne \emptyset$ and
for every  $s' \in \widehat{\Phi}(\alpha,s) $,  $\delta' \subseteq s'$.
This means that a sound approximation can be used for computing
conformant plans. In the rest of this section, we define a sound approximation of $\Phi$ and use if for
conformant planning. Because the program $\Pi(\mathcal{P},n)$ implements $\Phi$,
we define the approximation by describing a program $\Pi^a(\mathcal{P},n)$
approximating $\Phi$.

Let $a$ be an action and $\delta$ be a partial state. We say that $a$ is {\em executable} in $\delta$ if
$a$ occurs in an executability condition \eqref{exec} and each literal in the precondition of  the law holds in $\delta$.
A fluent literal $l$ is a {\em direct effect} (resp. a {\em possible direct effect}) of $a$ in $\delta$ if there exists a dynamic
causal law \eqref{dynamic} such that $\psi$ holds (resp. possibly holds) in $\delta$. Observe
that if $a$ is executable in $\delta$ then it  is executable in every state
$s \in comp(\delta)$. Furthermore, the direct effects of $a$ in $\delta$
are also the direct effects of $a$ in $s$, which in turn
are the possible direct effects of $a$ in $\delta$.

We next present the program $\Pi^a(\mathcal{P},n)$.  Atoms of $\Pi^a(\mathcal{P},n)$ are atoms of
$\Pi(\mathcal{P},n)$ and those formed by the following (sorted) predicate symbols:
\begin{itemize}
\item $de(l,T)$ is true if the fluent literal $l$ is a direct
effect of an action that occurs at the previous time step; and
\item $ph(l,T)$ is true if the fluent literal $l$ possibly
holds at time step $T$.
\end{itemize}
Similar to $holds(\varphi, T)$, we write
$\rho(\varphi,T) = \{ \rho(l,T) \mid  l \in \varphi \}$
and $\naf \rho(\varphi,T) = \{\naf \rho(l,T) \mid l \in \varphi \}$ for  $\rho \in \{holds, de, ph\}$.
The rules in  $\Pi^a(\mathcal{P},n)$  include most of the rules from
$\Pi(\mathcal{P},n)$, except for the inertial rules, which need to be changed. In addition,
it includes rules for reasoning about the direct
and possible effects of actions.

\begin{enumerate}
\item For each dynamic causal law \eqref{dynamic} in ${D}$,
$\Pi^a(\mathcal{P},n)$  contains the rule  \eqref{ir_dyn} and the next rule
\begin{eqnarray}
de(f, T+1)  & \leftarrow &  time(T),
    occ(a,T),  holds(\varphi,T) \label{dynamic_2}
\end{eqnarray}
%

This rule indicates that $f$ is a direct effect of the execution of $a$.
The possible effects of $a$ at $T$ are encoded using the  rule
\begin{equation}  \label{dynamic_3}
ph(f,T+1) \leftarrow time(T),
     occ(a,T), \naf holds(\overline{\varphi}, T) , \naf de(\overline{f},T+1)
\end{equation}
which says that $f$ might hold at $T+1$
if $a$ occurs at $T$ and the precondition $\varphi$
possibly holds at $T$.

\item For each static causal law \eqref{static} in ${D}$,
$\Pi^a(\mathcal{P},n)$  contains the rule \eqref{ir_sta} and the next rule
\begin{eqnarray}
ph(f,T)& {\leftarrow} & time(T), ph(\varphi,T) \label{static_3}
\end{eqnarray}
This rule propagates the possible holds relation between fluent literals.

\item The rule \eqref{ir_pos}  stating that $a$ can occur if its executability condition
is satisfied.
\item The inertial law is encoded as follows:
%
\begin{eqnarray}
ph(L,T+1) & \leftarrow & time(T), fluent(L), \naf holds(\neg {L},T),
    \naf de(\neg {L},T+1) \quad \quad   \label{iner_1} \\
ph(\neg L,T+1) & \leftarrow & time(T), fluent(L), \naf holds( {L},T),
    \naf de( {L},T+1) \label{iner_11} \\
holds(L,T) & \leftarrow & time(T), fluent(L),  \naf ph(\neg {L},T), T \ne 0. \label{iner_21} \\
holds(\neg L,T) & \leftarrow & time(T), fluent(L),  \naf ph({L},T), T \ne 0. \label{iner_2}
\end{eqnarray}
%
These rules capture the fact that  $L$ holds at  time moment $T > 0$
if its negation cannot possibly hold at $T$. Furthermore,
$L$ possibly holds at  time moment $T+1$
if its negation does not hold at $T$ and is not a direct effect of the action occurring at $T$.
These rules, when used in conjunction with the rules~\eqref{dynamic_2}-\eqref{dynamic_3},
compute the effects of the occurrence of action at time moment $T$.

\item $\Pi^a(\mathcal{P},n)$ also contains the rules of the form \eqref{ir_constraint},
\eqref{occ}, \eqref{constraint}, and the rule encoding the initial state and the goal
state as in $\Pi(\mathcal{P},n)$.

\end{enumerate}
It can be shown that if $\delta$ is a partial state and $x$ is an action executable in $\delta$
then the program $\Pi^a(\mathcal{P},1) \cup \{occ(x, 0)\}$, where $\mathcal{P} = \langle D, \delta, \emptyset \rangle$,
has a unique answer set $M$ and $\{f \mid holds(f, 1) \in M\}$ is a partial state of $D$.
For this reason, $\Pi^a(\mathcal{P},n)$ can be used to define a sound approximation
$\Phi^a$ of $\Phi$. The soundness of $\Phi^a$ is discussed in details by \citeN{tusogemo11a}.
This property  indicates that $\Pi^a(\mathcal{P},n)$ can be used as an ASP implementation of a conformant
planner. The planner \cpasp{}, as described by \citeN{tusogemo11a}, employs this implementation.

\begin{remark}
\begin{enumerate}

\item The key difference between \cpasp{} and \dlvk{} is the use of an approximation
semantics, which leads to the elimination of the security check in \cpasp{} and the use
of a single call to the answer set solver to find a solution.

\item \citeN{eifalepfpo03b} defined the notion of sound and complete security check
that can be used in the second step of \dlvk{} algorithm. They also identified
classes of planning problems in which different security checks are sound and
complete. The security check described in Subsection~\ref{subsection:dlvk}
is an adaptation of the check $\mathcal{SC}_1$ in the paper describing \dlvk{}. It is sound and complete for
domains called  \emph{{\tt false}-committed planning domains}. For consistent
action theories considered in this paper, $\mathcal{SC}_1$ is sound and complete. Observe
that this security check could be used together with the program $\Pi(\mathcal{P},n)$ in
the previous section to compute secure plans for non-deterministic domains.

\item Alternative representation approaches may facilitate the search of solutions in certain
domains. For example, as discussed by  \citeN{eifalepfpo03a}, the knowledge-state planning approach enables certain fluents to remain
open (i.e., as three-valued fluents), simplifying the state representation.
Actions enable to either
gain or forget knowledge of such fluents. For example, in the bomb-in-the-toilet domain, encoded in
\dlvk{}, this approach makes optimistic and secure plans equivalent.

\item \cpasp{}, the conformant planner using  $\Pi^a(\mathcal{P},n)$, performs well against logic-based
conformant planning systems (e.g., \dlvk{}).  \citeN{tusogemo11a} shows that
\cpasp{} and other logic-based systems are not as efficient and scalable in
common benchmarks  used by the planning community.
\cpasp{} also does not consider planning problem with disjunctive information. However,
their performance is superior in domains with static causal laws  \cite{sotugemo05b}.
In addition, logic-based planning systems can generate \emph{parallel plans,}
while  the existing heuristic search-based state-of-the-art conformant planning systems do not.

\item Approximation has its own price. Approximation-based planning systems are incomplete.
\cpasp{}, for example, cannot solve the problem
$\mathcal{P}^1_{inc} = \langle D^1_{inc}, \emptyset, f \rangle$ where
$D^1_{inc}$ consists of two dynamic laws:
$$\causes(a, f, \{g\}) \quad \quad \textnormal{ and } \quad\quad \causes(a, f, \{\neg g\}).$$
More specifically, $\Pi^a(\mathcal{P}^1_{inc},1)$ has no answer sets, while
$\mathcal{P}_{inc}$ has solution $\langle a \rangle$.

Similarly, \cpasp{} cannot solve the problem
$\mathcal{P}^2_{inc} = \langle D^2_{inc}, \emptyset, g \rangle$ where
$D^2_{inc}$ consists of the following laws:
\[
\causes(a, f, \emptyset) \quad \quad \caused(\{f,h\},g) \quad \quad \caused(\{f,\neg h\}, g)
\]
The main reason for the incompleteness of \cpasp{} is that it fails to reason by cases.
Syntactic conditions that guarantee that the proposed approximation is complete were proposed by  
\citeN{tusogemo11a} and \citeN{sontu06a}. Those authors also showed that the majority of  
planning benchmarks in the literature satisfy the conditions.  
The reason why  \cpasp{} cannot solve some of the benchmarks is
related to the presence of a disjunctive formulae in the  initial state.

\item Conformant planning using approximation is a successful approach
in dealing with incomplete information.  \citeN{trngsopo13a} have shown that,
with additional techniques to reduce the search space, such as goal splitting and
combination of {\tt one-of} clauses, approximation-based
planners perform exceptionally well compared to heuristic-based planning systems.

\item As with planning with complete information, ASP-based conformant planners such as \cpasp{}
do not include any heuristics (e.g., as those discussed in Subsection~\ref{subsection:heuristic}). This is a reason for the weak performance of \cpasp{} compared
to its search-based counterparts.

The second reason that greatly affects the performance of ASP-based planners is the need
for grounding before solving. In many benchmarks used by the planning community (see, e.g., the paper by
\citeN{trngsopo13a}  for details), the initial belief state of a small instance already contains $2^{10}$
states and the minimal plan length can easily reach 50. In most instances, grounding already fails.
We believe that, besides the use of heuristic, adapting techniques to reduce the search space
such as those proposed by \citeN{trngsopo13a}
and developing incremental grounding technique for ASP solver (e.g., the work by \cite{padoporo09a})
could help to scale up ASP-based conformant planners.

\item Various approximation semantics for action domains with static causal laws
have been defined \cite{sonbar01a,sotugemo05a,tusoba07a}. A discussion on their
strengths and weaknesses can be found in the paper by \citeN{tusogemo11a}.  A discussion of the performance of
\cpasp{}  against other planning systems is included in the next section.
\end{enumerate}
\end{remark}

\subsection{Context: Conformant Planning}

As with classical planning, several search-based conformant planners have been developed during the last three decades. Among them are  GPT \cite{bongef00a}, CGP \cite{smiwel98a},  CMBT \cite{cimrov00a},
 Conformant-FF (\cff) \cite{hofbra06a}, \kacmbp\  \cite{cirobe04a},  \pond\ \cite{brkasm06a},
\t0 \cite{palgef07a,palgef09a}, \T1 \cite{alrage11a}, \cpa \footnote{
   Different versions of  \cpa\
	have been developed. In this paper, whenever we refer to \cpa,  we mean \cpa(H), the version used in IPC 2008, \url{http://ippc-2008.loria.fr/wiki/index.php/Main_Page}.
} \cite{sotugemo05b,trngposo09a}, \cpls\ \cite{ngtrsopo11a}, \dnf\  \cite{toposo09a}, \cnf\  \cite{toposo10a}, \pip\ \cite{toposo10b},
\gclama\ \cite{ngtrsopo12a}, and
{\sc CPCES} \cite{grasca20a}. With the exception of CMBT, \kacmbp, \t0, and \gclama, the others planners are
forward search-based planners.


Differently from classical planning, a significant hurdle in the development of an efficient and scalable conformant planner is the size of the initial belief state and the size of the search space, which is double exponential in the size of the planning problem (see, e.g., the work by \cite{trngsopo13a} for a detailed discussion on this issue). Each of the aforementioned planners deals with this challenge in a different way.  Different representations of belief states are used in \cff, \dnf, \cnf, and \pip.
Specifically, \cff\ and \T1 make use of an implicit representation of a belief state as a sequence of actions from the initial state to the belief state.
\kacmbp, CMBP,  and \pond\ use a BDD-based representation \cite{bryant92a}.
\cpa\  approximates a belief state using a set of subsets of states ({\em partial states}).
\t0 and \gclama\ translate a conformant planning problem to a classical planning problem and
use  classical planners to compute solutions,
avoiding the need to deal with an explicit belief state representation.  Additional improvements
have been proposed in terms of heuristics
\cite{brkasm06a} and techniques   to reduce the size of
the initial belief state, such as the $\oneof$-combination technique
\cite{trngsopo13a}. Such a technique
is useful for planners employing an explicit disjunctive
representation of belief states, as in  \cpa\ \cite{trngsopo13a} and \dnf\
\cite{toposo09a}; a significant amount of work is
 required to apply this technique to other planners, due to
the different representations they use. Likewise, the
$\oneof$-relaxation technique proposed by \citeN{toposo10a} is  useful in \cnf\ but is
difficult to use in  other planners. Additional techniques proposed to improve
performance include  extensions of the {\sc GraphPlan} to deal with incomplete information, used in CGP,
 backward search algorithms, as in CMBT, landmarks, used in  \cpls,
 and sampling technique,  used in {\sc CPCES}.

SAT-based conformant planning is studied by several researchers \cite{cagita03a,palgef05a,rintanen99b}. The system
\cplan{}  \cite{cagita03a} has similarities  to \dlvk{}, in that it starts with a
translation of the planning problem into a SAT-problem, identifies a potential plan, and then validates the plan.
 \citeN{palgef05a} propose the system {\sc compile-project-sat}, which uses a single call to the SAT-solver to compute a conformant plan.
They show that the validity check can be done in linear time if the planning problem is encoded in a logically equivalent theory in deterministic decomposable negation normal form (d-DNNF).
As {\sc compile-project-sat} calls the SAT-solver only once, it is similar to \cpasp{}.
However, {\sc compile-project-sat} is complete, while \cpasp{} is not.
The system {\sc QBFPlan} by \citeN{rintanen99b} differs from \cplan{} and {\sc compile-project-sat} in that it translates the problem into a QBF-formula and uses  a QBF-solver to compute the solutions. A detailed comparison between these planning systems with \cpasp{}, directly or indirectly, can be found in the paper by \citeN{tusogemo11a}.

\section{Planning with Sensing Actions}
\label{section:sensing}

Conformant planning aims at addressing the completeness assumption of
the initial state in
classical planning but there are planning problems that do not admit
any
conformant plans as solution. The following example demonstrates
this issue.
\begin{example}
[From the work by \citeN{tusoba07a}]
\label{ex:conditional-plan}
Consider a security window with a lock that behaves
as follows. The window can be in one of the three
states {\em open, closed,}\footnote{
  The window is closed and unlocked.
}  or {\em locked.}\footnote{
  The window is closed and locked.
}
When the window is closed or open,
pushing it {\em up} or {\em down} will {\em open} or
{\em close} it, respectively.
When the window is not open, flipping the lock will
bring it to either the \emph{close} or \emph{locked} status.

Let us consider a security robot that needs to make sure that the window
is locked after 9pm. The robot has been told that
the window is not open (but whether it is locked or closed is unknown).

Intuitively, the robot can achieve its goal by performing
the following steps:
\begin{list}{}{\topsep=1pt \parsep=0pt \itemsep=1pt \leftmargin=12pt}
	\item[(1)] It checks the window
	to determine the window's status.
	\item[(2a)] If the window is in the \emph{closed} status,  
	the robot will lock the window;
	\item[(2b)] otherwise (i.e., the window is already in the \emph{locked} status)
 		the robot will not need to do anything.
 \end{list}
Observe that no sequence of actions
can achieve the goal from every possible initial situation. In other
words, {\em there exists no conformant plan} achieving the goal.
\hfill $\Diamond$
\end{example}

\subsection{Action Language $\mathcal{B}$ with Sensing Actions and Conditional Plans}

In order to solve the planning problem in Example~\ref{ex:conditional-plan},
sensing actions are necessary. Intuitively, the execution of a sensing action does
not change the world; instead, it changes  the knowledge of the action's performer. We
extend the language $\mathcal{B}$ with \emph{knowledge laws} of the following form:
\begin{eqnarray}
&  \determines(a,\theta) \label{knowledge}
\end{eqnarray}
where $\theta$ is a set of fluent literals. An action occurring in a knowledge law
is referred to as a \emph{sensing action.} The knowledge law states that the values of
the literals in $\theta$,
sometimes referred to as {\em sensed literals},
will be known after $a$ is executed. It is assumed that
the literals in $\theta$ are mutually exclusive, i.e.,
\begin{enumerate}
\item
for every pair of literals $g$ and $g'$ in $\theta$, $g \ne g'$,
the theory contains the static causal law
\[
\caused(\{g\}, \neg g')
\]
and

\item for every literal $g$ in $\theta$, the theory contains the
static causal law
\[
\caused(\{\neg g' \mid g' \in \theta \setminus \{ g \}\}, g).
\]
\end{enumerate}
We refer to this collection of static causal laws as
$\oneof(\theta)$.
We sometimes write
$
\determines(a,f)
$
as a shorthand  for
$
\determines(a,\{f,\neg f\}).
$

\begin{example}
\label{ex:conditional-plan-domain}
The planning problem instance $\mathcal{P}_{window} = (D_{window},\Gamma_{window},\Delta_{window})$
in Example \ref{ex:conditional-plan} can be represented as follows.
\[
D_{window} = \left \{
\begin{array}{lll}
\executable(push\_up, \{ closed\}) \\
\executable(push\_down, \{ open\}) \\
\executable(flip\_lock, \{ \neg open\}) \\
\\
\causes(push\_down, closed, \{\}) \\
\causes(push\_up, open,  \{\}) \\
\causes(flip\_lock, locked,  \{closed\}) \\
\causes(flip\_lock, closed,  \{locked\}) \\
\\
\oneof(\{open, locked, closed\})\\
\\
\determines(check, \{open,closed,locked\})
\end{array}
\right .
\]
\[
\Gamma_{window} \;\; = \; \left \{
\begin{array}{lll}
\initially(\neg open)
\end{array}
\right \}
\]
$$
\Delta_{window} = \{ locked \}
$$
\end{example}

It has been pointed out by several researchers 
\cite{warren76a,peosmi92a,prycol96a,levesque96a,lometa97a,sonbar01a,turner02a} that the notion of a plan
needs to be extended beyond a sequence
of actions, in order to allow conditional statements such as
{\bf if-then-else}, {\bf while-do}, or {\bf case-endcase}
to deal with incomplete information and sensing actions.
If we are interested in bounded-length plans, then the following
notion of \emph{conditional plans} is sufficient.

\begin{definition}[Conditional Plan]
\label{condplan}
\begin{enumerate}
\item  The empty plan, i.e., the plan
$\langle \rangle$ containing no actions, is a conditional plan.
\item
If $a$ is a non-sensing action and $p$ is a conditional plan then
$\langle a;p \rangle$ is a conditional plan.
\item\label{case-plan}
If $a$ is a sensing action with knowledge law (\ref{knowledge}),
where $\theta = \{ g_1, \dots, g_n \}$, and $p_j$'s
are conditional plans then $\langle a; \kcases(\{g_j \rightarrow p_j\}_{j=1}^{n}) \rangle$
is a conditional plan.
\item
Nothing else is a conditional plan.
\end{enumerate}
\end{definition}
Clearly, the notion of a conditional plan is more general than the notion of a plan as a sequence of actions.
We refer to the conditional plan
in Item~\ref{case-plan} of Definition~\ref{condplan} as a \emph{case-plan} and
the $g_j \rightarrow p_j$'s as its branches.
Under the above definition, the following are two possible conditional plans in
the domain $D_{window}$:
$$
p_1 = \langle push\_down;flip\_lock \rangle
$$
$$p_2 =
\left \langle
check; \;
\kcases \left (
\begin{array}{lll}
open & \rightarrow & [] \\
closed & \rightarrow & [flip\_lock] \\
locked & \rightarrow & []
\end{array}
\right ) \right \rangle
$$

\noindent
The semantics of the language $\mathcal{B}$ with knowledge laws also needs to be
extended to account for  sensing actions. Observe that, since it is possible that
the initial state of the planning problem is incomplete, we will continue using
the approximation $\Phi^a$ proposed in the previous section
as well as other notions, such as partial states, a fluent literal holds or possibly holds
in a partial state, etc. To reason about the effects of sensing actions in a domain $D$,
we define
\begin{equation} \label{closure-sensing}
\Phi^a(a,\delta) = \{Cl_{D}(\delta \cup \{g \}) \mid
g \in \theta \textnormal{ and }
Cl_{D}(\delta \cup \{g \}) \textnormal{ is consistent}\}
\end{equation}
where $a$ is a sensing action executable in $\delta$ and $\theta$ is the set of sensed literals of $a$. 
If $a$ is not executable in $\delta$ then $\Phi^a(a,\delta) = \bot$ (undefined).
The definition of $\Phi^a(a,\delta)$  for a non-sensing action is defined as in the previous section.
Intuitively, the execution of $a$ can result in several partial states,
in each of which exactly one sensed-literal in $\theta$ holds.

As an example, consider ${D}_{window}$ in Example \ref{ex:conditional-plan} and 
a partial state $\delta_1 = \{\neg open\}$. We have
\[
\begin{array}{rcl}
Cl_{D_{window}}(\delta_1 \cup \{open\}) & = &
\{ open, \neg open, closed, \neg closed, locked, \neg locked\} = \delta_{1,1} \\
Cl_{D_{window}}(\delta_1 \cup \{closed\}) & = &
\{ \neg open, closed, \neg locked \} = \delta_{1,2} \\
Cl_{D_{window}}(\delta_1 \cup \{locked\}) & =  &
\{ \neg open, \neg closed,  locked \} = \delta_{1,3}
\end{array}
\]
Among those, $\delta_{1,1}$ is inconsistent. Therefore,
we have $\Phi^a(check,\delta_1)=\{\delta_{1,2},\delta_{1,3}\}$.

The extended transition function $\widehat{\Phi^a}$ for computing the
result of the execution of a conditional plan is defined as follows.
Let $\alpha$ be a conditional plan and $\delta$ a partial state.
\begin{enumerate}
\item If $\alpha = \langle  \rangle $ then $\widehat{\Phi^a}(\alpha, \delta) = \delta$.

\item If $\alpha = \langle a; \beta \rangle$
and $a$ is a non-sensing action and $\beta$ is a conditional plan then
$\widehat{\Phi^a}(\alpha, \delta) =  \widehat{\Phi^a}(\beta,  \Phi^a(a, \delta))$.

\item
if $\alpha = \langle a; \kcases(\{g_j \rightarrow \alpha_j\}_{j=1}^{n}) \rangle$ where
$a$ is a sensing action with the sensed-literals
$\theta = \{ g_1, \dots, g_n \}$ and $\alpha_j$'s
are conditional plans then

\begin{enumerate}
\item $\widehat{\Phi^a}(\alpha, \delta) =  \widehat{\Phi^a}(\alpha_k,  \delta_k)$
if there exists $\delta_k \in \Phi^a(a, \delta)$ such that $\delta_k \models g_k$.

\item $\widehat{\Phi^a}(\alpha, \delta) =  \bot$ (undefined), otherwise.

\end{enumerate}

\item $\widehat{\Phi^a}(\alpha, \bot) =  \bot$ for every $\alpha$.
\end{enumerate}
%
Intuitively, the execution of a conditional plan progresses similarly to the execution of
an action sequence until it encounters a case-plan. In this case,
the sensing action is executed and the satisfaction of the formula in each branch is
evaluated. If one of the formulae holds in the state resulting after the execution of the sensing
action then the execution continues with the plan on that branch. Otherwise, the execution fails.

\subsection{ASP-Based Conditional Planning}\label{cond-planning}

Let us now describe an ASP-based conditional planner, called \cps{}, capable
of generating conditional plans. The  planner is a simple variation
of the one described by \citeN{tusoba07a}. As in the previous sections, we 
translate  a planning problem $\mathcal{P} = (D,\Gamma,\Delta)$ into
a logic program $\Pi_{h,w}(\mathcal{P})$ whose
answer sets represent solutions of $\mathcal{P}$. Before we present the rules
of $\Pi_{h,w}(\mathcal{P})$, let us provide the  intuition underlying the encoding.

\begin{figure}[bht]
\begin{center}
\includegraphics[width=.5\textwidth]{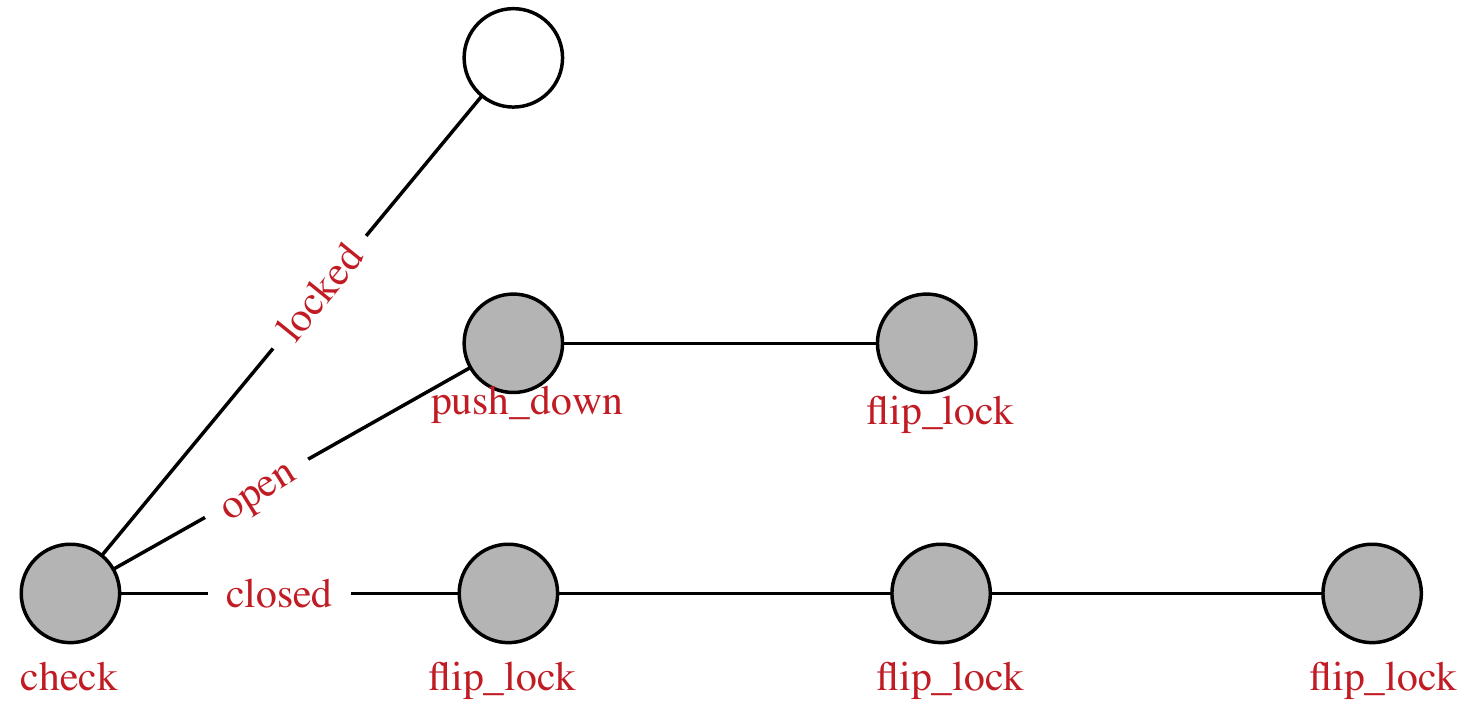}
\end{center}
\caption{A plan tree}
\label{tree}
\end{figure}

First, let us observe that each plan $\alpha$
(Definition \ref{condplan}) corresponds to a labeled plan
tree $T_\alpha$ defined as follows.
\begin{list}{$\bullet$}{\topsep=1pt \parsep=0pt \itemsep=1pt}
\item If $\alpha = \langle \: \rangle$   then $T_\alpha$ is a tree with a single node.
\item If $\alpha = \langle a \rangle$, where $a$ is a non-sensing action, then
$T_\alpha$ is a tree with a single node and this node is labeled with $a$.
\item If $\alpha = \langle a; \beta \rangle$, where $a$ is a non-sensing action and $\beta$ is
a non-empty plan, then $T_\alpha$ is a tree whose root is labeled with
$a$ and has only one subtree which is $T_\beta$. Furthermore, the
link between $a$ and $T_\beta$'s root is labeled with an empty string.
\item If $\alpha = \langle a;\kcases(\{g_j \rightarrow \alpha_j\}_{j=1}^n) \rangle$, where
$a$ is a sensing action, then $T_\alpha$ is
a tree whose root is labeled with $a$ and has $n$ subtrees
$\{T_{\alpha_j} \mid j \in \{1,\dots,n\}\}$.
For each $j$, the link from $a$ to the root
of $T_{\alpha_j}$ is labeled with $g_j$.
\end{list}
For example, the plan tree for the plan
\[
\alpha =
\left \langle check;
\kcases \left (
        \begin{array}{lll}
                locked & \rightarrow &  \langle \rangle; \\
                open & \rightarrow  & \langle push\_down;flip\_lock\rangle; \\
                closed & \rightarrow & \langle  flip\_lock; flip\_lock; flip\_lock \rangle
                \end{array}
         \right )
\right \rangle
\]
 is given in Figure~\ref{tree} (shaded nodes
indicate that there exists an action occurring
at those nodes, while white nodes indicate
that there is no action occurring at those nodes).

For a plan $p$, let $w(p)$ be the number of leaves of
$T_p$ and $h(p)$ be the number of nodes along the longest path
from the root to the leaves of $T_p$. $w(p)$ and $h(p)$
are called the {\em width} and {\em height} of $T_p$ respectively.
Suppose $w$ and $h$ are two integers that such that
$w(p) \le w$ and $h(p) \le h$.

\begin{figure}[bht]
\centering{\includegraphics[width=\textwidth]{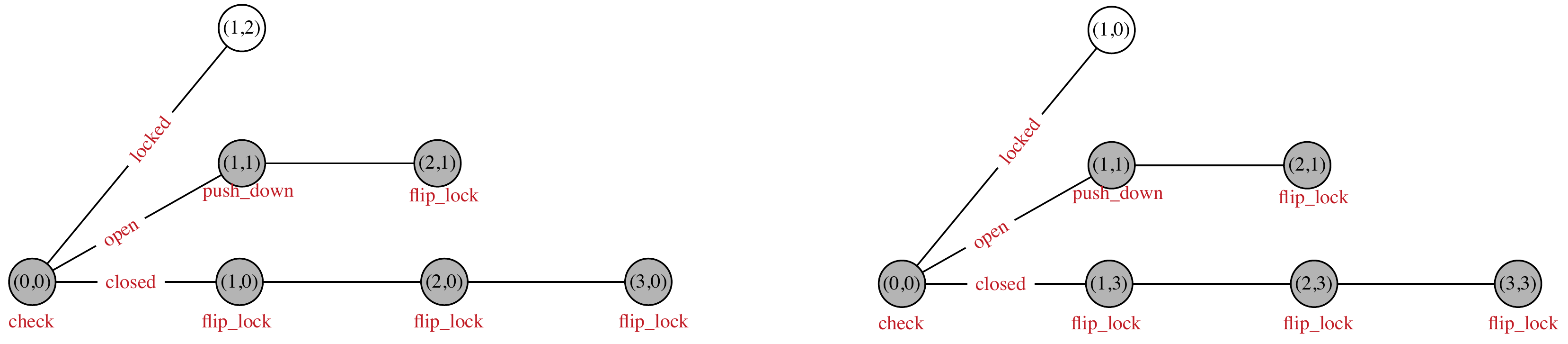}}
\caption{Possible mappings for the tree in Figure \ref{tree}}
\label{tree1}
\end{figure}

Let us denote the leaves of $T_p$ by $x_1,\ldots,x_{w(p)}$.
We map each node $y$ of $T_p$ to a pair of integers $n_y$ = ($t_y$,$p_y$),
where $t_y$, called the $t$-value of $y$, is the number of nodes along the path from
the root to $y$ minus 1, and $p_y$, called the $p$-value of $y$, is defined in the following way:
\begin{itemize}
\item For each leaf $x_i$ of $T_p$, $p_{x_i}$ is
an arbitrary integer between $0$ and $w$. Furthermore,
there exists a leaf $x$ such that $p_x = 0$, and
there exist no $i \ne j$ such that $p_{x_i} = p_{x_j}$.
\item For each interior node $y$ of $T_p$ with children
$y_1,\ldots,y_r$, $p_y = \min\{p_{y_1},\ldots,p_{y_r}\}$.
\end{itemize}
For instance, Figure \ref{tree1} shows some possible mappings
with $h=3$ and $w=3$ for the tree in Figure \ref{tree}.
It is easy to see that if $w(p) \le w$ and $h(p) \le h$ then
such a mapping always exists. Furthermore, from the construction
of $T_\alpha$, independently of how the leaves of $T_\alpha$ are numbered, we
have the following properties.
\begin{enumerate}
\item For every node $y$, $t_y \le h$ and $p_y \le w$.
\item For a node $y$, all of its children have the same $t$-value.
That is, if $y$ has $r$ children $y_1,\ldots,y_r$ then $t_{y_i} = t_{y_j}$
for every $1 \le i,j \le r$. Furthermore, the $p$-value of $y$
is the smallest one among the $p$-values of its children.
\item The root of $T_\alpha$ is always mapped to the pair $(0,0)$.
\end{enumerate}

The numbering schema of a plan tree provides a method for
generating a conditional plan on a two-dimensional coordinated system (or grid)
where the x- and y-axis correspond to the height and width of the plan tree,
and where $(0,0)$ is the initial state. Along a line of the same $y$-value is an action
sequence and the execution of a sensing action creates branches on different
lines, parallel to the $x$-axis.
For example, the execution of the $check$ action in the initial state of the plan tree in
Figure~\ref{tree1} creates three branches, to lines 0, 1, and 2. In the following, we use \emph{path}
to indicate the branch number and refer to a coordinate $(x,y)$ as a \emph{node}.

Let us next describe the rules in program $\Pi_{h,w}(\mathcal{P})$. Intuitively,
the program is similar to the program $\Pi^a(\mathcal{P},n)$ in that it implements
the approximation $\Phi^a$ and extends it to deal with sensing actions. Since a conditional
plan is two-dimensional, all  predicates $holds$, $ph$, $possible$,
$occ$, $de$, etc. need to extend with a third parameter. That is,
$holds(L,T,P)$---encoding that $L$ holds at node $(T,P)$ (the time step $T$ and the line number $P$ on the
 two-dimensional grid)---is used instead of $holds(L,T)$. In addition, $\Pi_{h,w}(\mathcal{P})$ uses
the following additional atoms and predicates.

\begin{itemize}
\item $path(0..w)$.
\item $sense(a,g)$ if $g$ is a sensed literal which belongs to $\theta$ in a knowledge law of the form \eqref{knowledge}.
\item $br(G,T,P,P_1)$ is true if there exists a branch from $(T,P)$
to $(T+1,P_1)$ labeled with $G$. 

For example, in Figure~\ref{tree1} (left),
we have $br(open,0,0,1)$, $br(closed,0,0,0)$, and $br(locked,0,1,2)$.

\item $used(T,P)$ is true if $(T,P)$ belongs to some extended branch of the plan tree.
This allows us to know which paths are used in the construction of the plan and allows us
 to check if the plan satisfies the goal. 
 
 In Figure \ref{tree1} (left),
we have $used(t,0)$ for $0 \le t \le h$, and
$used(t,1)$ and $used(t,2)$ for $1 \le t \le h$.
\end{itemize}
The rules of $\Pi_{h,w}(\mathcal{P})$ are divided into two groups.
The first group consists of rules from $\Pi^a(\mathcal{P},n)$
adapted to the two dimensional array for conditional planning.
The second group consists of rules for dealing with sensing actions.
We next describe the first group of rules\footnote{
   We omit $time(T), path(P)$ from the rules for brevity.
} in
$\Pi_{h,w}(\mathcal{P})$:
%
\begin{eqnarray}
holds(\Gamma,0,0)& \leftarrow & \label{r_1} \\
possible(a,T,P)& \leftarrow & holds(\varphi,T,P)  \label{r_2} \\
holds(f, T+1, P)  & \leftarrow &       occ(a,T, P),  holds(\varphi,T,P)  \label{r_30} \\
de(f, T+1, P)  & \leftarrow &       occ(a,T, P),  holds(\varphi,T,P)  \label{r_3} \\
ph(f,T+1, P) & \leftarrow &       occ(a,T, P),  \naf h(\overline{\varphi}, T, P) ,  \naf de(\overline{f},T+1,P) \quad \label{r_4} \\
ph(f,T,P)& {\leftarrow} &  ph(\varphi,T, P) \label{r_5} \\
holds(f,T,P)& \leftarrow &  holds(\varphi,T,P)  \label{r_6} \\
ph(L,T+1,P)& \leftarrow & fluent(L), \naf holds(\neg L,T), \naf de(\neg L,T,P) \label{r_7}\\
ph(\neg L,T+1,P)& \leftarrow & fluent(L), \naf holds(L,T), \naf de(L,T,P) \label{r_71}\\
holds(L,T+1,P)& \leftarrow & fluent(L),   \naf ph(\neg L,T,P) \label{r_81} \\
holds(\neg L,T+1,P)& \leftarrow & fluent(L),   \naf ph( L,T,P) \label{r_8} \\
1\{occ(X,T,P) : action(X)\} 1 & \leftarrow &     used(T,P), \naf goal(T,P)  \label{grr_1}\\
& \leftarrow &  occ(A,T,P), \naf possible(A,T,P)  \label{gr_2}
\end{eqnarray}
In the above rules, $L$ is a fluent literal, $T$ is a time moment in the range $[0,h-1]$,
and $P$ is in the range $[0,w]$. Rule \eqref{r_1} encodes the initial state. The rules
\eqref{r_2}--\eqref{r_8} are used for computing the effects of the occurrence of a non-sensing
action at the node $(T,P)$.  The rules
\eqref{grr_1} and \eqref{gr_2} are used for generating action occurrences,
similarly to the rules for generating action occurrences in the previous sections.
The difference is that the selection restricts the generation of action occurrences to nodes
marked as `used' (see below).

\begin{figure}[hb]
\centering{\includegraphics[width=\textwidth]{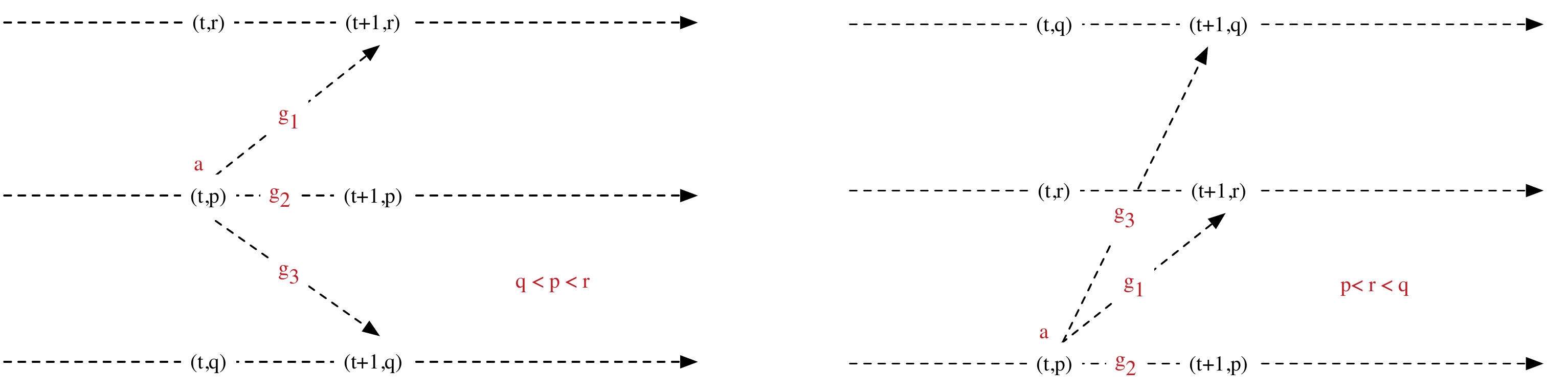}}
\caption{Sensing action $a$ that senses $\{g_1,g_2,g_3\}$ occurs at $(t,p)$ - disallowed (Left) vs. allowed (Right)}
\label{sensing-occ}
\end{figure}
The key distinction between $\Pi_{h,w}(\mathcal{P})$ and $\Pi^a(\mathcal{P},n)$ lies in the
rules for dealing with sensing actions. We next describe this set of rules.
\begin{itemize}

\item \emph{Rules for reasoning about the effect of sensing actions:}
For each knowledge law \eqref{knowledge} in $D$, $\Pi_{h,w}(\mathcal{P})$
contains the following rules:
\begin{eqnarray}
 1\{br(g,T,P,X){:}new\_br(P,X)\}1  & \leftarrow &
    occ(a,T,P), sense(a,g).  \label{rs_1} \\
& \leftarrow & occ(a,T,P), sense(a, g), \nonumber \\
                & &  \naf br(g,T,P,P) \label{rs_2} \\
& \leftarrow & occ(a,T,P), sense(a,g), \nonumber \\
                & & holds(g, T,P) \label{rs_3}  \\
new\_br(P,P_1) & \leftarrow &  P \le P_1  \label{rs_4}
\end{eqnarray}
When a sensing action occurs, it creates one branch for each of its sensed literals.
This is encoded in the rule~\eqref{rs_1}.  The constraint
\eqref{rs_2} makes sure that the current branch $P$ is continuing
if a sensing action occurs at $(T,P)$. The rule
\eqref{rs_3} is a constraint that
prevents a sensing action to occur if one of its sensed literals
 is already known. To simplify the selection of branches, rule
\eqref{rs_4} forces a new branch  at least at the same level as the current branch.
The intuition behinds these rules can be seen in Figure~\ref{sensing-occ}.

\item \emph{Inertia rules for sensing actions:}
This group of rules encodes the fact that the execution of
a sensing action does not change the world. However,
there is a one-to-one correspondence between the set of sensed
literals and the set of possible partial states.
\begin{eqnarray}
& \leftarrow & P_1 < P_2, P_2 < P, br(G_1,T,P_1,P), \nonumber \\
  & & br(G_2,T,P_2,P) \label{ri_1}  \\
& \leftarrow & P_1 \le P, G_1 \ne G_2, br(G_1,T,P_1,P), \nonumber \\
  & & br(G_2,T,P_1,P) \label{ri_2}  \\
& \leftarrow & P_1 <  P, br(G,T,P_1,P), used(T,P) \label{ri_3} \\
used(T+1,P) & \leftarrow &  P_1 < P, br(G,T,P_1,P)\label{ri_4} \\
holds(G,T+1,P) & \leftarrow & P_1 \le P, br(G,T,P_1,P)   \label{ri_5} \\
holds(L,T+1,P)& \leftarrow & P_1 < P, br(G,T,P_1,P),
holds(L,T,P_1) \label{ri_6} \\
used(0,0) & \leftarrow &   \label{ri_7} \\
used(T+1,P)& \leftarrow &  used(T,P) \label{ri_8}
\end{eqnarray}
The first three rules, together with rule \eqref{rs_4}, make sure that branches are separate from each other.
The next rule is used to mark a node as used if there is a branch in the plan that
reaches that node. This allows us to know which paths on
the grid are used in the construction of the plan and allows us
to check if the plan satisfies the goal (see rule
\eqref{gr_1}). The two rules \eqref{ri_5}--\eqref{ri_6}, along with rule \eqref{ri_4},
encode the possible partial state corresponding to the branch
denoted by literal $G$ after a sensing action is performed at $(T,P_1)$.
They indicate that the partial state at $(T+1,P)$ should contain $G$
(Rule \eqref{ri_5}) and literals that hold in $(T,P_1)$
(Rule \eqref{ri_6}). The last two rules mark nodes that have been used
in the construction of the conditional plan.

\item \emph{Goal representation:} Checking for goal satisfaction needs
to be done on all branches. This is encoded as follows.
\begin{eqnarray}
goal(T_1,P) & \leftarrow & holds(\Delta,T_1,P)  \label{gr_1}\\
goal(T_1,P) & \leftarrow & holds(L,T_1,P), holds(\neg L,T_1,P)  \label{r_13}\\
& \leftarrow & used(h+1,P), \naf goal(h+1,P) \label{r_14}
\end{eqnarray}
The first rule in this group says that the goal is satisfied at a node
if all of its subgoals are satisfied at that node.
The last rule guarantees that if a path $P$ is used in the construction
of a plan then the goal must be satisfied at the end of this path,
that is, at node $(h,P)$.
The second rule provides an
avenue to stop the generation of actions when an inconsistent state is
encountered---by declaring the goal reached. As discussed by \citeN{tusoba07a}, the
properties of the encoding of consistent action theories prevent this
method from generating plans leading to inconsistent states.
\end{itemize}

\begin{remark}

\begin{enumerate}
\item $\Pi_{h,w}(\mathcal{P})$ is slightly different from the program presented in the paper by \citeN{tusoba07a}
in that $\Phi^a$ is slightly different from the semantics used in that paper. By setting $w = 0$, this program is a
conformant planner. The experiments conducted by \citeN{tusoba07a} show that $\Pi_{h,w}(\mathcal{P})$ performs
reasonably well.

\item Extracting a conditional plan from an answer set $S$ of $\Pi_{h,w}(\mathcal{P})$
is not as simple as it is done in the previous sections because of the case-plan.
For any pair of integers $i$ and $k$ such that
$0 \le i \le h, 0 \le k \le w$, we define $p^k_i(S)$ as follows:
\[
p^k_i(S) = \left \{
\begin{array}{ll}
\langle \rangle & \textnormal{if } i = h  \textnormal{ or } occ(a,i,k) \not \in S
   \textnormal { for all } a   \\
\langle a;p^k_{i+1}(S) \rangle & \textnormal{if } occ(a,i,k) \in S \textnormal { and } \\
   & a  \textnormal{ is a non-sensing action}\\
\langle a;\kcases(\{g_j \rightarrow p_{i+1}^{k_j}(S)\}^n_{j=1}) \rangle &
\textnormal{if } occ(a,i,k) \in S, \\
& a  \textnormal{ is a sensing action},  \textnormal {and }\\
& br(g_j,i,k,k_j) \in S
\textnormal{ for } 1 \le j \le n
\end{array}
\right.
\]
Intuitively, $p^k_i(S)$ is the conditional plan whose corresponding
tree is rooted at node $(i,k)$ on the grid $h \times w$. Therefore, $p^0_0(S)$
is a solution to $\mathcal{P}$.

\item The semantics of $\mathcal{B}$ with knowledge laws
does not prevent a sensing action to occur when some of its sensed literals is known. It is easy to see that in this case,
the branching, enforced by rule~\eqref{rs_1}, is unnecessary. Rule~\eqref{rs_3} disallows such
redundant action occurrences. It is shown by \citeN{tusoba07a} that any
solution of $\mathcal{P} = \langle D, \Gamma, \Delta \rangle$
can be reduced to an equivalent plan without redundant occurrences of sensing actions
which can be found by  $\Pi_{h,w}(\mathcal{P})$.

\item  Because  the execution of a sensing action creates multiple branches and
some of them might be inconsistent (Eq.~\eqref{closure-sensing}),
rule (\ref{r_13}) prevents any action to occur at node $(T,P)$ when
the partial state at $(T,P)$ is inconsistent. To mark that the path ends
at this node, we say that the goal is achieved. \citeN{tusoba07a} showed that
for a consistent planning problem,  any solution generated by
$\Pi_{h,w}(\mathcal{P})$ corresponds to a correct solution.

\item The comparison between ASP-based systems, like \cpasp{} and $\Pi_{h,w}(\mathcal{P})$, and
conformant planning or conditional planning systems, such as  CMBP \cite{cimrov00a}, \dlvk\ \cite{eifalepfpo03b},
\cplan\ \cite{cagita03a},  \cff\ \cite{brahof04a},
\kacmbp\ \cite{cirobe04a}, $t_0$ \cite{palgef07a}, and POND \cite{brkasm06a}, has been
presented in the papers by \citeN{tusoba07a} and \citeyear{tusogemo11a}. The comparison  shows that ASP-based planning systems
perform much better than other systems in domains with static causal laws.

\item $\Pi_{h,w}(\mathcal{P})$, similar to \cpasp{}, makes a single call to the ASP solver to compute a
conditional plan. This is possible because of it uses an approximation semantics that reduces the complexity
of conditional planning, for polynomially-bounded plan, to NP-complete. Otherwise, this would not be possible
because conditional planning for polynomially-bounded length plan is PSPACE-complete \cite{turner02a}. Naturally,
as with \cpasp{}, this also implies that $\Pi_{h,w}(\mathcal{P})$ is incomplete.
\end{enumerate}

\end{remark}

\subsection{Context: Conditional Planning}

As we mentioned earlier, the need for plans with conditionals and/or loops has been identified very earlier on  
by \citeN{warren76a}, who developed Warplan-C, a Prolog program that can generate conditional plans and programs given the problem specification.
Warplan-C has only 66 clauses and is conjectured to be complete.
The system was developed at the same time as other non-linear planning systems, such as Noah by \citeN{sacerdoti74a}.
These earlier systems do not deal with sensing actions.
Other systems that generate plans with \textbf{if-then} statements and can prepare for contingencies are {\sc Cassandra} \cite{prycol96a} and {\sc CNLP} \cite{peosmi92a}.
These two systems extend partial order planning algorithms for computing conditional plans.

{\sc XII} \cite{goetwe96a} and {\sc PUCCINI} \cite{golden98a} are two  systems that employ partial order planning to generate conditional plans  for problems with incomplete information and  can  deal with  sensing actions.
SGP \cite{weansm98a} and POND \cite{brkasm06a} are conditional planners that work with sensing actions. These systems extend the planning graph algorithm \cite{blufur97a} to deal with sensing actions.
The main difference between SGP and POND is that the former searches solutions within the planning graph, whereas the latter uses it as a means of computing the heuristic function.

CoPlaS, developed by \citeN{lobo98a}, is a regression Prolog-based planner that uses a high-level action description language, similar to the language 
$\mathcal{B}_K$ described in this section, to represent and reason about effects of actions, including sensing actions.
\citeN{nieeitver07} introduced $\mathcal{K}_c$, an extension of language $\mathcal{K}$, to deal with sensing actions and compute conditional plans as defined in this section using \dlvk. 
\citeN{thielscher00a} presented FLUX, a constraint logic programming based planner, which
is capable of generating and verifying conditional
plans.
{\sc QBFPlan} is another conditional planner, based on a QBF theorem prover, is described
 in the paper by  \citeN{rintanen99b}.  This system, however, does not consider sensing actions.

Research in developing conditional planners, however, has not attracted as much attention compared to other types of planning
domains in recent years.
Rather, the focus has been on synthesizing \emph{controllers} or \emph{reactive modules} which exhibit specific behaviors in different environments 
 \cite{amgilomuru20a,cabemc19a,cabamumc18a,trebel20a}.
This is similar to the effort of generating programs satisfying a specification as discussed earlier (e.g., the work by \citeN{warren76a}) or attempts to compute \emph{policies} (see, e.g., the book by \citeN{bellman57a}) for Markov Decision Processes (MDP) or Partially Observable Markov Decision Processes (POMDP).
To the best of our knowledge, little attention has been paid to this research direction within the ASP community.
We present this as a challenge to ASP in the last section of the paper.

\section{Planning with Preferences}\label{sec:prefs}

The previous  sections analyze answer set planning with the focus on solving different classes of planning
problems, such as planning with complete information, incomplete information, and sensing actions.
In this section, we present  another extension of the planning problem, by illustrating
the use of answer set planning in \emph{planning with preferences.}

The problem of planning with preferences arises in situations where the user not only wants a
plan to achieve a goal, but has specific preferences or biases about  the plan. This situation is
common when the  space of possible plans for a goal is dense, i.e., finding ``a'' plan is not difficult, but
many of the plans may have features which are undesirable to the user.
This type of situations is very common in practical planning problems.
\begin{example}
\label{pp:exp1}
Traveling from one place to another is a frequently considered problem (e.g., a traveler,
a transportation vehicle, an
autonomous vehicle).
A planning problem  in the travel domain can be represented by the following elements:

\begin{itemize}
\item a set of fluents of the form
$at(\ell)$, where $\ell$ denotes a location,
such as {\em home, school, neighbor, airport, etc.};

\item an initial location $\ell_i$;

\item a destination location $\ell_f$; and

\item a set of actions of the form $method(\ell_1,\ell_2)$
where $\ell_1$ and $\ell_2$ are two distinct locations and
$method$ is one of the available transportation methods, such
as {\em drive, walk, ride\_train, bus, taxi, fly, bike, etc.}
The problem may include  conditions that restrict the
applicability of actions in certain situations. For example,
one can ride a taxi only if the taxi has been called, which can
be done only if one  has some money; one can fly from one place
to another if one has a ticket; etc.
\end{itemize}
Problems in this domain are often rich in solutions
because of the large number of actions which can be used
in the construction of a plan. For this reason, a user looking  for a solution to
a problem often considers some additional features, or
\emph{personal preferences}, in selecting a plan. For example, the user might
be biased in terms of
the distance to travel using a transportation method, the overall cost, the
time to destination, the comfort of a vehicle, etc. However, a user would accept a plan
that does not satisfy her preferences if she has  no other choice.
\hfill$\Diamond$
\end{example}
Preferences can come in different shapes and forms. The most common
types of preferences are:
\begin{itemize}
\item \emph{Preferences about a state:} the user prefers to be in
a state $s$ that satisfies a property $\phi$ rather than
a state $s'$ that does not satisfy it, in case
both lead to the satisfaction of the goal; for example, being in a 5-star hotel is preferable
to being in a 1-star hotel, if the distance to the conference site is the same;
\item \emph{Preferences about an action:} the user prefers
to  perform (or avoid) an action $a$, whenever it is feasible and it
allows the goal to be achieved; for example, one might prefer to
walk to destination whenever possible;
\item \emph{Preferences about a trajectory:} the user prefers a
trajectory that satisfies a certain property $\psi$
over those that do not satisfy this property; for example, one might
prefer plans that do not involve traveling through Los Angeles
during peak traffic hours;
\item \emph{Multi-dimensional preferences:} the user has a
\emph{set} of preferences, with an ordering
among them. A plan satisfying a more
favorable preference is given priority over
those that satisfy less favorable preferences; for example,
plans that minimize time to destination might be preferable to
plans minimizing cost.
\end{itemize}
\citeN{sonpon06a} propose a general method for integrating diverse classes of
preferences into
answer set planning. Their approach is articulated in two
components:
\begin{itemize}
\item \emph{Development of a preference specification language}: this language
supports the specification of different types of preferences; its  semantics should enable the definition
of  a partial order among possible solutions of the planning problem.

\item \emph{Implementation}: the implementation proposed by \citeN{sonpon06a}
maps preference formulae to mathematical formulae, representing the \emph{weight} of
each formula,
and makes use of the $\mathtt{maximize}$ statement in answer set programming to optimize
the solution to the planning problem.
The original paper defines rules for computing the weights of preference formulae in ASP.
\end{itemize}
We next introduce the preference specification language proposed by \citeN{sonpon06a}.

\subsection{A Preference Specification Language}

The proposed preference specification language addresses the
description of  three classes of preferences: \emph{basic desires},
\emph{atomic preferences}, and \emph{general preferences}.

For a planning problem $\mathcal{P} = \langle D, \Gamma, \Delta \rangle$,
a basic desire can be in of the following possible forms:
\begin{list}{}{\topsep=1pt \parsep=0pt \itemsep=1pt}
	\item[{\bf (a)}] a \emph{state formula}, which is a
fluent formula $\varphi$ or a formula of the form $occ(a)$ for some action $a$,
or
	\item[{\bf (b)}] a \emph{goal  preference}, of the form
$\textbf{goal}(\varphi)$, where $\varphi$ is a fluent formula.
\end{list}

\paragraph{Basic desire formula.} A basic desire formula is a formula built over basic desires,
the traditional propositional operators ($\wedge$, $\vee$, and $\neg$), and the
modalities $\next$, $\always$, $\eventually$, and $\until$. The BNF for the basic
desire formulae is
\[
\psi \stackrel{def}{=} \varphi \mid \psi_1 \wedge \psi_2 \mid \psi_1 \vee \psi_2 \mid
				\neg \psi_1 \mid \next(\psi_1) \mid \until(\psi_1,\psi_2) \mid \always(\psi_1) \mid \eventually(\psi),
\]
where $\varphi$ represents a state formula or a goal preference and $\psi$, $\psi_1$, or $\psi_2$ are
basic desire formulae.

Intuitively, a basic desire formula specifies a property that a user
would like to see satisfied by the provided plan.
For example, to express the fact that a user would like to take the taxi or
the bus to go to school, we can write:
\[
\eventually(\:occ(bus(home,school)) \vee occ(taxi(home,school))\:).
\]
If the user's desire is not to call a taxi,  we can write
\[
\always(\:\neg occ(call\_taxi(home))\:).
\]
If the user's desire is not to see any taxi around his
home, we can use the basic desire formula
\[
\always(\:\neg available\_taxi(home)\:).
\]
Note that these encodings have different consequences---the
last formula  prevents taxis to be present, independently from whether
the taxi has been called.

The following are several basic desire formulae that
are often of interest to users.

\begin{itemize}
\item \emph{Strong Desire}:
	For two   formulae $\varphi_1$ and $\varphi_2$,
	 $\varphi_1 < \varphi_2$ denotes
	$\varphi_1 \wedge \neg \varphi_2$.

\item \emph{Weak Desire}:
	For two   formulae $\varphi_1$ and $\varphi_2$,
 	$\:\:\varphi_1 <^w \varphi_2$ denotes
	$\varphi_1 \vee \neg \varphi_2$.

\item \emph{Enabled Desire}:
	For two actions $a_1$ and $a_2$, 	$a_1 <^e a_2$ stands for  \\
	$(executable(a_1) \wedge executable(a_2)) \Rightarrow
		(occ(a_1) < occ(a_2))$
	where \\
	 $executable(a) = \bigwedge_{l \in \varphi} l $ if $\executable(a, \varphi) \in D$.

\item \emph{Action Class Desire}: For actions with the same set of parameters and
effects such as {\em drive} or {\em walk}, we write
$drive <^e walk$ to denote the desire \\
$
\bigvee_{l_1, l_2 \in S, \: l_1 \ne l_2} (drive(l_1, l_2) <^e walk(l_1, l_2))
$
where $S$ is a set of pre-defined locations. Intuitively, this preference states
that we prefer to drive rather than to walk between locations belonging to the
set $S$. For example, if $S = \{home, school\}$ then this preference says
that we prefer to drive from home to school and vice versa.

\end{itemize}

\paragraph{Atomic preference.}
Basic desire formulae are expressive enough to describe
a significant portion of preferences that frequently occur in real-world
domains. It is also often the case that the user may provide a variety of
desires, and some desires are stronger than others; knowledge of such
biases about desires becomes important when it is not possible to
concurrently satisfy all the provided desires. In this situation,
an ordering among the desires is introduced.
An \emph{atomic preference formula} is 	a formula
of the form
\[
 \varphi_1 \lhd \varphi_2 \lhd \cdots \lhd \varphi_n
 \]
where $\varphi_1, \dots, \varphi_n$ are basic desire formulae.
The atomic preference formula states that   trajectories that satisfy $\varphi_1$ are preferable
to those that satisfy $\varphi_2$, etc.

Let us consider again the travel domain.
Besides {\em time} and {\em cost}, users often have their preferences
based on the level of comfort and/or safety of  the available
transportation methods. These preferences can be represented by the
formulae\footnote{The notation $\alpha <^e \beta <^e \gamma$ is a syntactic
	sugar for $\alpha <^e \beta \wedge \beta <^e \gamma$.}
\[
\mathit{cost} = \always(\mathit{walk} <^e \mathit{bus} <^e \mathit{drive}  <^e \mathit{flight})
\]
and
\[
\mathit{time} = \always(\mathit{flight} <^e \mathit{drive} <^e \mathit{bus}  <^e \mathit{walk})
\]
and
\[
\mathit{comfort} = \always(\mathit{flight} <^e (drive \vee bus) <^e walk)
\]
and
\[
\mathit{safety}  = \always(walk <^e \mathit{flight} <^e (drive \vee bus)).
\]
These four desires can be combined to produce the following
two atomic preferences
\[
\Psi^t_1 = \mathit{comfort} \lhd \mathit{safety} \:\:\: \textnormal{ and } \:\:\:
\Psi^t_2 = cost \lhd time.
\]
Intuitively, $\Psi^t_1$ is a comparison between level of comfort and
safety, while $\Psi^t_2$ is a comparison between affordability
and duration.

\paragraph{General preference formulae.} Suppose that a user
would like to travel as comfortably \emph{and} as cheaply as possible.
Such a preference can be viewed as a multi-dimensional preference,
which cannot be easily captured using
atomic preferences or basic desires. \emph{General preference formulae} support
the
representation of such multi-dimensional preferences.

Formally, a  general preference formula is a formula satisfying one of the
following conditions:
\begin{itemize}
\item An atomic preference $\Psi$ is a general preference;
\item If $\Psi_1, \Psi_2$ are general preferences, then
		$\Psi_1 \& \Psi_2$, $\Psi_1 \mid \Psi_2$,
                 and
		$!\: \Psi_1$ are general preferences;
\item If $\Psi_1, \Psi_2, \dots, \Psi_k$ is a
        collection of general preferences, then
	$\Psi_1 \lhd \Psi_2 \lhd \cdots \lhd \Psi_k$ is
	 a general preference.
\end{itemize}
In the above definition, the operators $\&, \mid, !$ are used to express
different ways to combine preferences. For example, the preference $\Psi^t_1 \& \Psi^t_2$
indicates that the user prefers trajectories that are most preferred according to both $\Psi^t_1$
and $\Psi^t_2$; $\Psi^t_1 \mid \Psi^t_2$ states that, among trajectories with the same cost,
the user prefers trajectories that are most comfortable or vice versa. A detailed discussion of
general preferences can be found in the paper by \citeN{sonpon06a}.

\paragraph{Semantics.}
In order to define the semantics of the preference language, we need to start from
describing  whether
a trajectory $\alpha = s_0a_0\ldots a_{n-1} s_n$ satisfies a basic desire formula.
We write
$\alpha \models \varphi$ to denote that $\alpha$ satisfies a basic desire $\varphi$.
The definition of $\models$ is straightforward, and we report here only some of the cases
(the complete definition can be found in the paper by \citeN{sonpon06a}), where $\alpha[i] = s_i a_i \ldots a_{n-1}s_n$.
\begin{itemize}
	\item $\alpha \models occ(a)$ if $a_0=a$;
	\item $\alpha \models \ell$ if $s_0\models \ell$ and $\ell$ is a fluent;
	\item $\alpha \models \always(\varphi)$ if for all $0\leq i < n$ we have $\alpha[i]\models\varphi$;
	\item $\alpha \models \next(\varphi)$ if $\alpha[1] \models \varphi$.
\end{itemize}

The satisfaction of desires is  used to define two relations between trajectories, one expressing
preference between trajectories and one capturing the fact that two trajectories are
indistinguishable,  denoted by $\prec_\Psi$ and $\approx_{\Psi}$, respectively,
where $\Psi$ is a preference formula. For two
trajectories $\alpha$ and $\beta$,
\begin{enumerate}
\item if $\Psi$ is a basic desire then $\alpha \prec_\Psi \beta$ ($\alpha$ is more preferred than $\beta$ with respect to $\Psi$)
if $\alpha \models \Psi$ and $\beta \not\models \Psi$;
$\alpha \approx_{\Psi} \beta$ denotes that $\alpha \models \Psi$ iff $\beta \models \Psi$.

\item if $\Psi$ is an atomic preference $\varphi_1 \lhd \varphi_2 \lhd \cdots \lhd \varphi_n$ then
$\alpha \prec_{\Psi} \beta$ if $\exists (1 \leq i \leq n)$ such that
\begin{enumerate}
\item $\forall (1 \leq j < i)$ we have that $\alpha \approx_{\varphi_j} \beta$,
and
\item  $\alpha \prec_{\varphi_i} \beta$.
\end{enumerate}

$\alpha \approx_{\Psi} \beta$ denotes that $\alpha  \approx_{\varphi_j} \Psi$
for every $j=1,\ldots,n$.

\item if $\Psi$ is a general preference and has the form $\Psi = \Psi_1 \lhd \cdots \lhd \Psi_k$ then
$\alpha \prec_\Psi \beta$ is defined similar to the second case. Otherwise,
$\alpha \prec_\Psi \beta$
\begin{enumerate}
\item if $\Psi = \Psi_1 \& \Psi_2$ and
	$\alpha \prec_{\Psi_1} \beta$ and
	$\alpha \prec_{\Psi_2} \beta$
\item if $\Psi = \Psi_1 \mid \Psi_2$ and  (\emph{i})
		$\alpha \prec_{\Psi_1} \beta$ and
		$\alpha \approx_{\Psi_2} \beta$; or (\emph{ii})
   $\alpha \prec_{\Psi_2} \beta$ and
		$\alpha \approx_{\Psi_1} \beta$; or (\emph{iii})
  $\alpha \prec_{\Psi_1} \beta$ and
		$\alpha \prec_{\Psi_2} \beta$.

\item if $\Psi = \:! \Psi_1$ and
        $\beta \prec_{\Psi_1} \alpha$.

\end{enumerate}

In all cases, $\alpha  \approx_{\Psi} \beta$ iff $\alpha  \approx_{\Psi'} \beta$ where $\Psi'$ is a component of $\Psi$.
\end{enumerate}
The following proposition holds \cite{sonpon06a}.
\begin{proposition}
\label{pref:prop1}
$\prec_\Psi$ is a partial order and $\approx_{\Psi}$ is an equivalent relation.
\end{proposition}
The above proposition shows that maximal elements with respect to $\prec_\Psi$ exist, i.e.,
most preferred trajectories exist.

\subsection{Implementation: Computing Preferred Plans}

Given a planning problem $\mathcal{P}$ and a preference formula $\Psi$, a preferred trajectory can be computed
using the following steps:
\begin{enumerate}
\item Use one of the programs, denoted by $\mathit{Plan}(\mathcal{P},n)$,
introduced in Sections~\ref{asp-sec}--\ref{section:conformant} to compute a potential
solution $\alpha$ for $\mathcal{P}$ ;
\item Associate to $\alpha$ a number, which represents the degree of satisfaction of $\alpha$ with respect to $\Psi$; and
\item Use the $\#maximize$ statement of \clingo{} to compute a most preferred trajectory.
\end{enumerate}
This process requires an appropriate encoding of $\Psi$. This is usually achieved by
converting $\Psi$ to a canonical form, which provides a convenient way to translate $\Psi$ into a set of facts
with predefined predicates. For example, a basic desire formula can be encoded using the predicates \emph{and},
\emph{or}, $\neg$, \emph{occ}, $\next$, $\until$, $\eventually$, $\always$, and \emph{final} (stands for \emph{goal}, to avoid
confusion with the \emph{goal} predicate defined in the previous sections). This translation can be done
using a script (e.g., \citeN{sonpon06a} presented a Prolog program for such translation).

An atomic preference can be represented using an ordered set, consisting of a declaration
$atomic(id)$, indicating that $id$ is an atomic preference, and a set of atoms of the form
$member(id, \mathit{formula}, order)$, where $id$, \emph{formula}, and \emph{order}
represent the atomic preference identifier, the formula, and the order of the formula in $id$, respectively.
Finally, a general preference can be encoded using the predicates $\&$, $\mid$, and $!$, and the basic
representations of basic desires and atomic preferences.
In the following, we use $\Pi(\Psi)$ to denote the set of facts encoding $\Psi$.

Since the first task in computing a most preferred trajectory is checking whether or not a trajectory satisfies
a basic desire, we need to add to $\mathit{Plan}(\mathcal{P},n)$ rules for this purpose. This task
can be achieved by a set of domain-independent rules $\Pi_{\mathit{pref}}$  defining
 the predicate $holds(sat(\varphi), t)$, which states
that the basic desire formula $\varphi$ is satisfied by the trajectory $s_t a_t \ldots a_{n-1} s_n$.
$\Pi_{\mathit{pref}}$ contains the following groups of rules:
\begin{itemize}
\item  {Rules for checking the satisfaction of a propositional formula at a time step:} such as
\[
\begin{array}{lcl}
holds(sat(F), T) &\leftarrow & fluent(F), holds(F,T)\\
holds(sat(and(F,G)), T) &\leftarrow& holds(sat(F), T), holds(sat(G),T)
\end{array}\]

\item  {Rules for checking the satisfaction of a temporal formula at a time step:} such as
$$holds(sat(\next(F)), T) \leftarrow holds(sat(F), T+1)$$

\item  {Rules for checking the occurrence of an action:}
$$holds(sat(occ(A)), T) \leftarrow occ(A,T)$$

\item  {Rules for checking the satisfaction of a goal formula:}
$$holds(sat(final(F)), 0) \leftarrow holds(sat(F), n)$$

\end{itemize}
The following proposition holds.
\begin{proposition}
\label{pref:prop2}
An answer set $S$ of $\mathit{Plan}(\mathcal{P},n) \cup \Pi(\varphi) \cup \Pi_{\mathit{pref}}$
contains $holds(sat(\varphi), 0)$ if and only if the trajectory corresponds to $S$ satisfies $\varphi$.
\end{proposition}
The above proposition shows that $\mathit{Plan}(\mathcal{P},n) \cup \Pi(\varphi) \cup \Pi_{\mathit{pref}}$ can be used
to compute a most preferred trajectory with respect to a basic desire formula. To do so,
we only need to tell \clingo{} that an answer set containing $holds(sat(\varphi), 0)$ is preferred.

 To compute a most preferred plan with respect to a general preference or an atomic formula,
\citeN{sonpon06a} proposed a set of rules that assigns a number to a formula
and then use the $\#maximize$ statement of \clingo{} to select a most preferred trajectory.
The proposed rules were developed at the time the answer set solvers did provide only limited
capability to work with numbers. For this reason, we omit the detail about these rules here.
Features provided in more recent versions of answer set solvers, such as
multiple
optimization statements and weighted tuples, are likely to enable a simpler and
more efficient implementation. For example,
we could translate an atomic preference
\[
\varphi_1 \lhd \varphi_2 \ldots \lhd \varphi_n
\]
to a statement
\[
\#maximize \{1@n : holds(sat(\varphi_1), 0); \ldots ; n@1 : holds(sat(\varphi_n), 0)\}
\]
as a part of $\Pi(\Psi)$.

\begin{remark}
\begin{enumerate}
\item The encoding of $\Pi(\Psi)$ presented in this paper does not employ any advanced features of answer set programming, which were introduced to simplify the encoding of preferences such as the framework for preferences specification \asprin{},  introduced
by \citeN{brderosc15a} and \citeyear{brderosc15b}. 
Analogously, a more elegant encoding of  preference formulae can be achieved using other extensions of answer
set programming focused on rankings of answer sets, such as logic programming with ordered disjunctions
\cite{brewka02a}. This encoding, however, cannot be used with \clingo{} or \dlv{} because none of these solvers supports 
ordered disjunctions.

%
%

\item The discussion in this section focuses on expressing preferences
over  trajectories, i.e., sequences of actions and states.
     It can be extended to conditional plans and used with the planner in Section~\ref{section:sensing}.
     This extension can, for example,  define preferences over branches in a conditional plan.

%
%
%

\end{enumerate}
\end{remark}

\subsection{Context: Planning with Preferences}

Planning with preferences has attracted a lot of attention by the planning community.
An excellent survey of several systems developed before 2008 and their strengths and weaknesses  can be found in the paper by \citeN{baimci08a}.
Preferences have also been included in  extensions of PDDL, such as PDDL3 \cite{gerlon05a}.
The majority of systems described in the survey employ PDDL3 where preferences are, ultimately, described by a numeric value.
As such, most of the planning with preferences systems in the literature can be characterized as \emph{cost optimal}, where the cost of actions plays  a key role in deciding the preference of a solution.
Hierarchical task planning is also frequently used in these systems.
Representative systems in this direction are  HPlan-P \cite{sobamc09a}, {\sc LPRPG-P} \cite{colcol11a}, and {\sc PGPlanner} \cite{daodisdorona19a}.
{\sc ChoPlan}, developed by \citeN{bipibe19a}, is a  system which encodes a PDDL3 planning problem as a \emph{multi-attribute utility theory} and a heuristic based on Choquet integrals to derive solutions.

{\sc satplan(P)}, developed by \citeN{giumar07a}, shows that SAT-based planning is also competitive with other planning paradigms.
\citeN{giumar11a} present a  survey of SAT-based planning with preferences.
In recent years, SMT-based planning has become more popular than SAT-based planning,
thanks to the expressiveness of SMT compared to SAT and the availability of efficient SMT solvers. SMT-based planning, which can work with numeric variables, provides an excellent way to deal with preferences \cite{scrahath16a}.
It is worth observing that ASP-based planning with action costs has been considered  earlier by \citeN{eifalepfpo03c} and more recently   by \citeN{khyalelist14a}.

\section{Planning and Diagnosis}\label{sec:diag}
While planning and diagnosis are often considered two separate and independent tasks,
some researchers have suggested that ties exist between them, to the point that it is
possible to reduce diagnostic reasoning to planning. In this section,
we present this view, and specifically focus on the approach from \citeN{bargel00a} and \citeN{balgel03b}, under which planning tasks and diagnostic tasks share (a) the same domain representation and (b) the same core reasoning algorithms.

In this section, the term \emph{diagnosis} describes a type of reasoning task in which an agent identifies and interprets discrepancies between the  domain's expected behavior
and the domain's actual/observed behavior. Consider the following example.
\begin{example}[From the paper by \citeN{balgel03b}]\label{diag:ex1}
Consider the
analog circuit $\cal {AC}$ from Figure \ref{diag:fig.output}.

\begin{figure}[htbp]
\begin{center}
\includegraphics{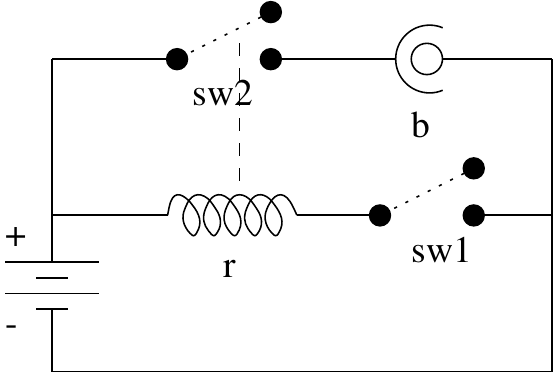}
\caption{Analog circuit $\cal{AC}$}\label{diag:fig.output}
\end{center}
\end{figure}

We assume that switches $sw_1$ and $sw_2$ are mechanical components that
cannot be damaged.  Relay {\em r} is a magnetic coil.
If not damaged, it is activated when $sw_1$ is closed,
causing $sw_2$ to  close.  Undamaged
bulb $b$ emits light if $sw_2$ is closed.
For simplicity of presentation
we consider an agent capable of performing only one action,
$close(sw_1)$. The environment can be represented
by two damaging exogenous\footnote{By {\em exogenous} actions
we mean actions performed by the agent's environment. This includes natural
events as well as actions performed by other agents.}
actions:
$brk$, which causes $b$ to become faulty, and
$srg$, which damages $r$ and also $b$
assuming that $b$ is not protected.
Suppose that the agent operating this device is given a goal
of lighting the bulb. He realizes that this can be achieved
by closing the first switch, performs the operation,
and discovers that the bulb is not lit. The domain's behavior does not match the agent's expectations.
 The agent needs to determine the reason for this state of affairs and  ways to correct the problem.
\hfill $\Diamond$
\end{example}

In the following, we focus on {\em non-intrusive} and {\em observable}
domains, in which the agent's environment does not
\emph{normally} interfere with his work and the agent
\emph{normally} observes all of the domain occurrences of exogenous actions.
The agent is, however, aware that
these assumptions can be contradicted by
observations. The agent is ready to observe and to take into
account occasional occurrences of exogenous actions that alter the behavior of the environment.
Moreover, discrepancies between expectations and observations
may force the agent to conclude that additional exogenous actions have occurred, but remained unobserved.

To model the domain, let us introduce a finite set of \emph{components} {\bf C}, disjoint from {\bf A} and {\bf F} previously introduced. Let us also assume the existence of a set {\bf F$_0$} $\subseteq$ {\bf F} of \emph{observable fluents} (i.e., fluents that can be directly observed by the agent), such that $ab(c) \in$ {\bf F$_0$} for every component of {\bf C}. Fluent $ab(c)$ intuitively indicates that the $c$ is ``faulty.'' Let us point out  that the use of the relation $ab$ in diagnosis dated back to the work by \citeN{reiter87b}.
The set {\bf A} is further partitioned into two
disjoint sets: {\bf A$_s$}, corresponding to \emph{agent actions}, and {\bf A$_e$} consisting of \emph{exogenous actions}. Additionally,
exogenous and agent actions are allowed to occur concurrently. With respect to the formalization methodology introduced in Section \ref{sub:languageB}, this is achieved by {\bf (1)} introducing the notion of \emph{compound action}, i.e., a set of agent and exogenous actions, {\bf (2)} redefining $\Phi(a,s)$ so that $a$ is a compound action, and
{\bf (3)} extending the notion of trajectory to be a sequence $s_0 a_0 \ldots a_{k-1} s_k$ of states and compound actions.

A core principle of this approach is that the \emph{discrepancies between agent's expectations and observations are explained in terms of occurrences of unobserved exogenous actions}. The observed behavior of the domain is represented by a particular trajectory, referred to as the {\em actual trajectory}.

A {\em diagnostic domain} is a pair
$\langle D, W \rangle$ where $D$ is a domain and $W$ is the domain's actual trajectory.

Information about the behavior of the domain up to a certain step $n$ is captured by the {\em recorded history} $H_n$, i.e. a set of {\em observations} of the form:
\begin{enumerate}
\item $obs(l,t)$  -- meaning that fluent literal
$l$ was observed to be true at step $t$;
\item
$hpd(a,t)$ -- stating that action $a \in$ {\bf A} was observed to happen at time $t$.
\end{enumerate}
The link between diagnostic domain and recorded history is established by the following:

Consider a diagnostic domain $\langle D, W \rangle$ with
$W = s^{w}_0a^{w}_0\ldots a^{w}_{n-1}s^{w}_n$, and let $H_n$ be a recorded history up to step  $n$.
\begin{enumerate}
\item
A trajectory $s_0a_0 \ldots a_{n-1}s_n$ is a {\em model} of $H_n$ if
for any $0 \leq t \leq n$
\begin{enumerate}
\item $a_t = \{a : hpd(a,t) \in H_n\}$;
\item  if $obs(l,t) \in H_n$ then $l \in s_t$.
\end{enumerate}
\item  $H_n$ is {\em consistent} if it has a model.
\item  $H_n$ is {\em sound} if, for any
$l$, $a$, and $t$, if $obs(l,t), hpd(a,t) \in H_n$ then
$l \in s^{w}_t$ and $a \in a^{w}_t$.
\item A fluent literal $l$ {\em holds} in a model $M$ of $H_n$ at time $t \leq n $ ($M \models holds(l,t)$) if $l \in s_t$; $H_n$ {\em entails} $h(l,t)$($H_n \models holds(l,t)$) if, for every model $M$ of $H_n$, $M \models holds(l,t)$.
\end{enumerate}
Note also that a recorded history may be
consistent, but not sound -- which is the case if the recorded history is incompatible with the actual trajectory.

Example \ref{diag:ex1} can thus be formalized as:
\[
\begin{array}{l}
\mbox{\% Fluents:}\\
fluent(active(r)) \leftarrow \hspace*{0.5in}fluent(on(b)) \leftarrow \hspace*{0.5in}  fluent(prot(b)) \leftarrow \\
fluent(closed(sw_1)) \leftarrow \hspace*{0.33in} fluent(closed(sw_2)) \leftarrow \\
fluent(ab(r)) \leftarrow \hspace*{0.72in}  fluent(ab(b)) \leftarrow\\
\mbox{\% Agent Actions:}\\
a\_act(close(sw_1)) \leftarrow\\
\mbox{\%Exogenous Actions}\\
x\_act(brk) \leftarrow \hspace*{0.9in}  x\_act(srg) \leftarrow
\end{array}
\]
Note the use of relations $a\_act$ and $x\_act$ to distinguish agent actions and exogenous actions.
The laws describing the normal and abnormal/malfunctioning behavior of the domain are:
\[
D_{\mathcal{AC}} = \left\{
\begin{array}{lll}
\mbox{\% normal } && \mbox{\% abnormal} \\
\causes(close(sw_1),closed(sw_1),\emptyset) && \causes(brk,ab(b),\emptyset)\\
\caused(\{closed(sw_1), \neg ab(r)\}, active(r))  && \causes(srg, ab(r),\emptyset)\\
\caused(\{active(r)\}, closed(sw_2))  && \causes(srg, ab(b),\{\neg prot(b)\})\\
\caused(\{closed(sw_2), \neg ab(b)\}, on(b))  && \caused(\{ab(b)\}, \neg on(b))\\
\caused(\{\neg closed(sw_2)\}, \neg on(b))  && \caused(\{ab(r)\}, \neg active(r))\\
\executable(close(sw_1), \{\neg closed(sw_1)\})
\end{array}
\right.
\]
Now consider a recorded history:
\[
H_1 =
\left\{
\begin{array}{l}
hpd(close(sw_1),0)\\
obs(\neg closed(sw_1),0)\\
obs(\neg closed(sw_2),0)\\
obs(\neg ab(b),0)\\
obs(\neg ab(r),0)\\
obs(prot(b),0)
\end{array}
\right.
\]
One can check that $s_0 \  close(sw_1) \  s_1$ is
the only model of $H_1$, where $s_0$ is the state depicted in Figure \ref{diag:fig.output}. Additionally, $H_1 \models holds(on(b),1)$.

Next, we formalize the key notions of diagnosis. Let $\delta=\langle D, W \rangle$ be a diagnostic domain. A \emph{configuration} is a pair
\begin{equation}\label{sympt}
{\cal S} = \langle H_n,O^{m}_n\rangle
\end{equation}
where $H_n$ is the recorded history up to step $n$
and $O^{m}_n$
is a set of observations between steps $n$ and $m \geq n$.
Leveraging this notion, we can now define a \emph{symptom} as a configuration $\langle H_n,O^{m}_n\rangle$ such that $H_{n}$ is consistent and $H_{n} \cup O^{m}_n$ is not.

Once a symptom has been identified, the next step of the diagnostic process aims at finding its possible reasons. Specifically, a \emph{diagnostic explanation} $E$ of symptom ${\cal S} = \langle H_n,O^{m}_n\rangle$ is defined as a set
\begin{equation}\label{exp}
E \subseteq \{hpd(a_i,t) : 0 \leq t < n \mbox{ and } a_i \in \mbox{\bf A}_e\},
\end{equation}
such that $H_{n} \cup O^{m}_n \cup E$ is consistent.

\subsection{Diagnostic Reasoning as Answer Set Planning}

As we said earlier, answer set planning can be used to determine whether a diagnosis is needed and for computing diagnostic explanations. Next, we introduce
an ASP-based program for these purposes.
Consider a domain $D$ whose behavior up to step $n$ is described by recorded history $H_n$. Reasoning about $D$ and $H_n$ can be accomplished by a translation to a logic program $\Pi(D, H_n)$ that follows the approach outlined in Section \ref{asp-sec}. Focusing for simplicity on the direct encoding, $\Pi(D, H_n)$ consists of:
\begin{itemize}
\item
the domain dependent rules from Section \ref{subdep};
\item
the rules related to inertia and consistency of states (\ref{inertial_1})-(\ref{constraint});
\item
domain independent rule establishing the relationship between observations and the basic relations of $\Pi$:
\begin{eqnarray}
occ(A,T) \leftarrow hpd(A,T) \\
holds(L,0) \leftarrow obs(L,0)
\end{eqnarray}
\item
the \emph{reality check axiom}, i.e. a rule ensuring that in any answer set the agent's expectations match the available observations (variable $L$ ranges over fluent literals):
\begin{eqnarray}
\leftarrow obs(L,T), \naf holds(L,T)
\end{eqnarray}
\end{itemize}
The following  theorem  establishes an important relationship between
models of a recorded history and answer sets of the corresponding logic program.
\begin{theorem}\label{th1-diagnosis}
If the initial situation of $H_n$ is {\em complete}, i.e. for any fluent $f$, $H_n$ contains $obs(f,0)$ or $obs(\neg f,0)$, then $M$ is a model of $H_n$ iff $M$ is defined by some answer set of $\Pi(D, H_n)$.
\end{theorem}
The proof of the theorem is in two steps. First, one shows that the theorem holds for $n=1$, i.e., that for a history $H_1$ there is a one-to-one correspondence between the transitions of the form $s_0a_0s_1$ and the answer sets of $\Pi(D,H_1)$. Then, induction is leveraged to extend the correspondence to histories of arbitrary length. The complete proof can be found in the paper by \citeN{balgel03b}.

Next, we focus on identifying the need for diagnosis. Given a domain $D$ and ${\cal S} = \langle H_n,O^{m}_n\rangle$, we introduce:
\begin{equation}\label{e3}
\mathit{TEST}({\cal S})= \Pi(D, H_n) \cup O^m_{n}.
\end{equation}
The following corollary forms the basis of this approach to diagnosis.
\begin{corollary}\label{c1}
Let ${\cal S} = \langle H_n,O^{m}_n\rangle$ where $H_n$
is consistent. A configuration ${\cal S}$ is a symptom iff $\mathit{TEST}({\cal S})$ has no answer set.
\end{corollary}
Once a symptom has been identified, diagnostic explanations can be found by means of the answer sets of the \emph{diagnostic program}
\begin{equation}\label{e4}
\Pi_d({\cal S})= \mathit{TEST}({\cal S}) \cup
\{\ \ \ 1 \{ occ(A,T) : x\_act(A) \} 1 \leftarrow time(T), T < n.\ \ \ \}
\end{equation}
Specifically, every answer set $X$ of $\Pi_d({\cal S})$ encodes a diagnostic
explanation
\[
E=\{hpd(a,t) \,\,|\,\, occ(a,t) \in X \land a \in \mbox{{\bf A}$_e$} \}.
\]
Note that the choice rule shown in (\ref{e4}) is simply a restriction of (\ref{occ}) to exogenous actions. As a result, $\Pi_d({\cal S})$ can be viewed as a variant of the translation $\Pi(\mathcal{P},n)$ of a planning problem, where planning occurs over the past $0..n\!-\!1$ time steps and over exogenous actions only, and the goal states are described by the observations from ${\cal S}$.
\begin{example}\label{diag:ex1a}
Consider the domain from Example \ref{diag:ex1}. According to $H_1$ initially switches $sw_1$ and $sw_2$ are open,
all circuit components are ok, $sw_1$ is
closed by the agent, and {\em b} is protected. The expectation is that
$b$ will be {\em on} at 1.
Suppose that, instead, the agent observes that at time 1
bulb {\em b} is {\em off}, i.e.
$O_1 = \{obs(\neg on(b),1) \}$.
$\mathit{TEST}({\cal S}_0)$, where ${\cal S}_0 = \langle H_1,O_1\rangle$, has no answer sets and thus, by
Corollary \ref{c1}, ${\cal S}_0$ is indeed a symptom.
The diagnostic explanations of ${\cal S}_0$ can be found by computing the answer sets of $\Pi_d(\mathcal{S})$. Specifically, there are three diagnostic explanations:
\[
\begin{array}{l}
E_1 = \{occ(brk,0)\} \\ 
E_2 = \{occ(srg,0)\} \\ 
E_3 = \{occ(brk,0), occ(srg,0)\}
\end{array}
\]
\end{example}

\begin{remark}
\begin{enumerate}
        \item Other interpretations of the relationship between agent and environment
        are possible, yielding substantial differences in the overall approach to diagnosis.
        The interested reader is referred to the paper by \citeN{bamcso00a}.
        \item In contrast to the approach by \citeN{bamcso00a}, the approach presented in this survey assumes that a recorded history is
        consistent only if observations about fluents can be explained
        without assuming the occurrence of actions not recorded in $H_n$.
	\item In the paper by \citeN{balgel03b}, the formalization of diagnostic reasoning presented here is extended to incorporate an account of the agent's interaction with the domain in order to collect physical evidence that confirms or refutes the diagnostic explanations computed. This is accomplished by introducing the notions of candidate diagnosis and of diagnosis.

        \item Theorem \ref{th1-diagnosis} is similar to the result from the paper by \citeN{turner97a}, which
        deals with a different language and uses the definitions by \citeN{mcctur95a}.
        If the initial situation of $H_n$ is incomplete, one can adopt techniques
        discussed elsewhere in this paper or the \emph{awareness axioms} by \citeN{balgel03b}.
	\item As discussed in the paper by \citeN{balgel03b}, the diagnostic process may not always lead to a unique solution. In those cases, the agent may need to perform further actions, such as repairing or replacing components, and observe their outcomes.  \citeN{balgel03b} provided a specialized algorithm to achieve this. An alternative, and potentially more general, option consists in leveraging conditional planning techniques (see Section \ref{cond-planning}), i.e., by creating a conditional plan that determines the true diagnosis as proposed by
\citeN{bamcso00a}.
\end{enumerate}
\end{remark}

\subsection{Context: Planning and Diagnosis}




Classical diagnosis such as the foundational work by \citeN{reiter87b} aimed at providing a formal answer to the question of ``\emph{what is wrong with a system given its (current failed) state}.''
Central to classical diagnosis is the use of the model of the system to reason about failures.
Earlier formalizations considered a single state of the system and are often referred to as \emph{model based diagnosis}, which is summarized by \citeN{dekkur03a}.
Later formalization such as the proposal by \citeN{fepiwomakl20a} considers a finite trace of states, taking into consideration the transitions at different time points in need of a diagnosis.
This is closedly related to \emph{dynamic diagnosis}, as described in this paper, which has been considered in the literature  \cite{bamcso00a,balapoza99a,corthi94a,mcilraith97a,thielscher97a,thcojekr96a,wilnay96a}.

The close relationship between model based diagnosis and satisfiability led to several methods for computing diagnosis using satisfiability such as the method proposed by \citeN{graanb13a}, \citeN{mestkaco14a}, or   described in several publications on diagnosing sequential circuits (e.g., by \citeN{fepiwomakl20a}). In a recent work by \citeN{wotawa20a}, answer set programming has been used in the context of model-based diagnosis.


\section{Planning in Multi-Agent Environment}
\label{sec:mas}

The formalizations presented in the previous sections can be also extended to deal with various problems in multi-agent environments (MAE). In these problems, the planning (or reasoning) activity can be carried out either by one system (a.k.a. \emph{centralized planning}) or multiple systems (a.k.a. \emph{distributed planning}). In the following subsections, we discuss the use of ASP in these settings.

\subsection{Centralized Multi-Agent Planning} \label{sub:cmap}

We will start with the \emph{Multi-Agent Path Finding (MAPF)} problem which appears
in a variety of application domains,  such as
autonomous aircraft towing vehicles \cite{mopalumamakuko16a}, autonomous warehouse
systems~\cite{wudamo08a}, office robots ~\cite{vebicoro15a}, and video games~\cite{silver05a}.

A MAPF problem is defined by a tuple $\mathcal{M} = (R, (V,E), s, d)$ where

\begin{itemize}
\item $R$ is a set of robots,

\item $(V,E)$ is an undirected graph, with the set of vertices $V$ and the set of edges $E$,

\item $s$ is an injective function from $R$ to $V$, $s(r)$ denotes the starting location of robot $r$, and

\item $d$ is an injective function from $R$ to $V$, $d(r)$ denotes the destination of robot $r$.
\end{itemize}
Robots can move from one vertex to one of its neighbors  in one step. A collision occurs
when two robots move to the same location (vertex collision) or traveling on the same edge (edge collision).
The goal is to find a collision-free plan for all robots to reach their destinations. Optimal plans,
with the minimal number of steps, are  often preferred.

\begin{figure}[htbp]
\centering
\includegraphics[width=.45\textwidth]{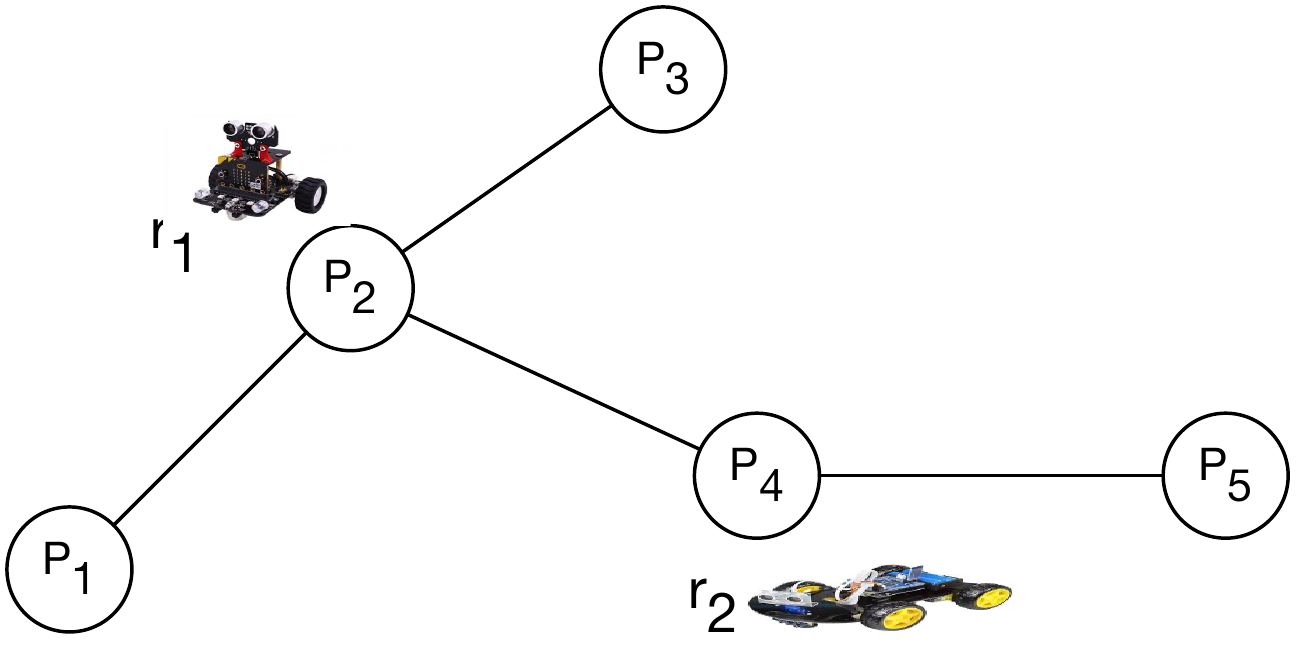}
\caption{A Multi-Agent Path Finding Problem}
\label{fig:mas}
\end{figure}

A simple MAPF problem is depicted in Figure~\ref{fig:mas}. In this problem,
we have two robots $r_1$ and $r_2$ on a graph with five vertices $p_1, \ldots, p_5$. Initially,
$r_1$ is at $p_2$ and  $r_2$ is at $p_4$. The goal consists of moving
robot $r_1$ to location  $p_5$ and robot $r_2$ to location $p_3$.

It is easy to see that a MAPF  $\mathcal{M} = (R, (V,E), s, d)$  can be represented by
\begin{itemize}

\item a set $\{\mathcal{P}_r \mid r \in R\}$ of path-planning problems for the robots in $R$, where for each $r \in R$,
$\mathcal{P}_r = \langle D_r, at(r, s(r)), at(r, d(r)) \rangle$ is a planning problem with
\[
D_r = \left \{
\begin{array}{lll}
\causes(move(r,l,l'), at(r,l'), \{at(r,l)\}) && \mbox{ for } (l,l') \in E \\
\caused(\{at(r,l)\}, \neg at(r,l')) \}  && \mbox{ for } l \ne l', l,l' \in V \\
\end{array}
\right.
\]

\item the set of constraints representing the collisions.
\end{itemize}
%
%
ASP-based solutions of MAPF problems have been proposed for both
action-based as well as state-based encodings. In the context of this survey,
we will focus on an action-based encoding.
Consider a MAPF problem $\mathcal{M}= (R, (V,E), s, d)$.
The program developed in Section~\ref{asp-sec} for a single-agent planning problem can be used
to develop a program solving $\mathcal{M}$, denoted by $\Pi(\mathcal{M},n)$, as follows.
$\Pi(\mathcal{M},n)$ contains the following groups of rules:

\begin{enumerate}

\item the set of atoms $\{agent(r) \mid r \in R\}$ encoding the robots;

\item the collection of the rules from $\Pi(\mathcal{P}_r, n)$; and

\item the constraints to avoid collisions:
\begin{eqnarray}
& \leftarrow & agent(R), agent(R'), R \ne R' ,  \label{v-collision} \\
&& holds(at(R, V), T), holds(at(R', V), T) \nonumber \\
& \leftarrow & agent(R), agent(R'), R \ne R' ,  \label{e-collision} \\
& & holds(at(R, V), T), holds(at(R', V'), T), \nonumber \\
& & holds(at(R, V'), T+1), holds(at(R', V), T+1) \nonumber
\end{eqnarray}
\end{enumerate}

Rule \eqref{v-collision} prevents two robots to be at the same location at the same time,
while rule \eqref{e-collision} guarantees that edge-collisions will not occur.  These two constraints
and the correctness of $\Pi(\mathcal{P}_r,n)$ imply that
$\Pi(\mathcal{M},n)$ computes solutions of length $n$ for the MAPF problem $\mathcal{M}$.

The proposed method for solving MAPF problems can be generalized to multi-agent planning
as considered by the multi-agent community in the setting discussed by  \citeN{durfee99a}. We assume that
a multi-agent planning problem $\mathcal{M}$ for a set of agents $R$ is specified by a pair $(\mathcal{P}_{r \in R}, \mathcal{C})$
where $\mathcal{P}_r = \langle D_r, \Delta_r, \Gamma_r \rangle $ is the planning problem for agent $r$ and $\mathcal{C}$ is a set of
global constraints. For simplicity of the presentation, we will assume that

\begin{itemize}
\item the agents in $R$ share the same set of fluents and, wherever needed, parameterized with the names of the agents; for example,
if an agent is carrying something then $carrying(r, o)$ will be used instead of $carrying(o)$ as in a single-agent domain;

\item the actions in the domain in $D_r$ are parameterized with the agent's name $r$, e.g., we will use
the action  $move(r,l,l')$ instead of the traditional encoding
 $move(l,l')$;

\item the constraints in $\mathcal{C}$ are of the form
\begin{equation}\label{eq:exec_m}
\executable(sa, \varphi)
\end{equation}
where $sa$ is a set of actions in $\bigcup_{r \in R} D_r$ and $\varphi$ is a  set of literals. This
can be used to represent parallel actions, non-concurrent actions, etc.
\end{itemize}

For a  multi-agent planning problem $\mathcal{M} = (\mathcal{P}_{r \in R}, \mathcal{C})$
where $\mathcal{P}_r = \langle D_r, \Delta_r, \Gamma_r \rangle$, the program  $\Pi(\mathcal{M},n)$
that computes solution for $\mathcal{M}$ consists of

\begin{enumerate}

\item the set of agent declarations, $agent(r)$ for $r \in R$;

\item the collection\footnote{
   We keep only one instance of the domain-independent rules.
} of the rules from $\Pi(\mathcal{P}_r, n)$ with the following modifications:

\begin{itemize}
\item the action specification of the form $action(a)$ is replaced by $action(r,a)$;

\item the action generation rule is replaced by
\[
1 \{occ(A, T) : action(R, A)\} 1 \leftarrow time(T), agent(R)
\]
\end{itemize}
where, without the loss of generality, we assume that every $D_r$ contains the action
\emph{noop} that is always executable and has no effect;

\item for each constraint of form \eqref{eq:exec_m}, we create a new
``collective'' action named $sa_{id}$, add $action(sa_{id})$ to the set of actions in $\mathcal{M}$ (and make the
\eqref{eq:exec_m} its executability condition, translated into ASP as for any other action)
and, for each $a \in sa$, the following rule is added to $\Pi(\mathcal{P}_r, n)$:
\begin{eqnarray}
occ(a, T) & \leftarrow & occ(sa_{id}, T) \label{occ:set-member}
\end{eqnarray}
\end{enumerate}

\subsection{Distributed Planning}
\label{sec:mas-disptributed}

A main drawback of centralized planning is that it cannot exploit the structural organization of agents (e.g., hierarchical organization of agents) in the planning process. Distributed planning has been proposed as an alternative to centralized planning that aims at exploiting the independence between agents and/or groups of agents. We discuss distributed planning in two settings: fully collaborative agents and partially-cooperative or non-cooperative agents.

\subsubsection{Fully Collaborative Agents}

When agents are fully collaborative, a possible way to exploit structural relationships between agents is to allow each group of agents to plan for itself (e.g., using  the planning system described in Section~\ref{asp-sec}
and then employ a centralized post-planning process (a.k.a. the \emph{controller/scheduler}) to create the joint plan for all agents. The controller takes the output of these planners---individual plans---and merges them into an overall plan. One of the main tasks of the controller is to resolve conflicts between individual plans. This issue arises because individual groups plan without knowledge of other groups (e.g., robot $r_1$ does not know the location of robot $r_2$). When the controller is unable to resolve all possible conflicts, the controller will identify plans that need to be changed and request different individual plans from specific individual groups.

Any implementation of distributed planning requires some communication capabilities between the controller and the individual planning systems. For this reason, a client-server architecture is often employed in the implementation of distributed planning. A client plans for an individual group of agents and the server is responsible for merging the individual plans from all groups. Although specialized parallel ASP solvers exist (e.g., the systems discussed in the papers by \citeN{lepon05a} and \citeN{scscgekakasc09a}), there has been no attempt to use parallel ASP solvers in distributed planning. Rather, distributed planning using ASP has been implemented using a combination of Prolog and ASP, where communication between server and clients is achieved through Prolog-based message passing, and planning is done using ASP (e.g., the system described in the paper by \citeN{sopong09a}).

Observe that the task of resolving conflicts is not straightforward  and can require multiple iterations with individual planner(s) before the  controller can create a joint plan. Consider again the two robots in Figure~\ref{fig:mas}. If they are to generate their own plans, then the first set of individual solutions can be
{
\begin{equation}\label{ma:sol2}
{
\begin{array}{l}
occ(move(r_1, p_2, p_4), 0), occ(move(r_1, p_4, p_5), 1)
\end{array}
}
\end{equation}
}
and
{
\begin{equation}\label{ma:sol3}
{
\begin{array}{l}
occ(move(r_2, p_4, p_2), 0),  occ(move(r_2, p_2, l_3), 1)
\end{array}
}
\end{equation}
}

A parallel execution of these two plans will result in
a violation of the constraint stating that two robots cannot be at the
same location at the same time. One can see that the controller needs to
insert a few actions into both plans (e.g., $r_1$ must move to either $l_1$
or $l_3$ before moving to $l_4$).

Let $\mathcal{M}$ be a multi-agent planning problem and $P_{\{r \in R\}}$
be the plans received by the controller. The feasibility of merging
these plans into a single plan for all agents can be checked using ASP.
Let $\pi_n$ be the program obtained from $\Pi(\mathcal{M},n)$ (described in  Subsection~\ref{sub:cmap})
by adding to $\Pi(\mathcal{M},n)$
\begin{itemize}
\item   the set of action occurrences in  $P_{\{r \in R\}}$, i.e.,
\[
\bigcup_{\{r \in R\}} \{occurs(a, t) \mid occ(a,t) \in P_r\}
\]
\item  for $r \in R$, rules mapping time steps from 0 to $n$ to time steps used in $P_r$ ($r \in R$),
\[
\begin{array}{l}
1 \{map(r, T, J) : time(J) \}  1 \leftarrow time(T), T< n, T \le \max_r \\
\leftarrow map(r, T, J), map(p, T', J'), T < T', J > J'
\end{array}
\]
where $\max_r$ is the maximal index in $P_r$. Intuitively, $map(r, i, j)$ indicates that the $i^{th}$ action in $P_r$
should occur in the $j^{th}$ position in the joint plan.
This mapping must conform to the order of action occurrences in $P_r$.

\item a rule ensuring that an atom $occurs(a, j) \in P_r$ must occur at the specified position:
\[
\leftarrow occ(a, t), map(r, t, j), \naf occurs(a, j).
\]
\end{itemize}
It can be checked that $\pi_4$ would generate an answer set consisting of
\[
\begin{array}{l}
occ(move(r_1, p_2, p_1), 0), occ(move(r_2, p_4, p_2), 0),\\
occ(move(r_2, p_2, l_3), 1),  occ(move(r_1, p_1, p_2), 1),\\
occ(move(r_1, p_2, p_1), 2), \\
occ(move(r_2, p_4, p_2), 3)
\end{array}
\]
which corresponds to the mapping $map(1, 0, 2), map(1, 1, 3), map(2, 0, 0), map(2, 1, 1)$ and is a successful merge of
the two plans in \eqref{ma:sol2}--\eqref{ma:sol3}.

Observe that the program $\pi_n$ might have no answer sets, which indicates that the merging of the plans $P_{\{r \in R\}}$ is unsuccessful.
For instance, $\pi_3$ has no answer set, i.e., the two plans in  \eqref{ma:sol2}--\eqref{ma:sol3} cannot be merged with less than four steps.

\subsubsection{Non/Partially-Collaborative Agents}
\label{subsubsec:mas-non-collaborative}

Centralized planning or distributed planning with an overall controller is most suitable in applications with collaborative (or non-competitive) agents such as the robots in the MAPF problems. In many applications, this assumption does not hold,  e.g., agents may need to withhold certain private information and thus do not want to share their information freely; or agents may be competitive and have conflicting goals. In these situations,
distributed planning as described in the previous sub-section is not applicable and planning will have to rely on a message passing architecture, e.g., via peer-to-peer communications. Furthermore, an online planning approach might be more appropriate. Next, we describe an ASP approach that is implemented centrally by \citeN{soposa09a} but could also be implemented distributedly.

In this approach, the planning process is interleaved with a negotiation process among agents. As an example, consider the robots in Figure~\ref{fig:mas} and assume that the robots can communicate with each other, but they cannot reveal their location. The following negotiation between $r_2$ and $r_1$ could take place:
\begin{itemize}
\item $r_2$ (to $r_1$):  ``can you ($r_1$) move out of $l_2$, $l_3$, and $l_4$?'' (because $r_2$ needs to make sure that it can move to location $l_2$ and $l_3$). This can be translated to the formula $\varphi_1 = \neg at(r_1, l_2) \wedge \neg at(r_1, l_3) \wedge \neg at(r_1, l_4)$ sent from $r_2$ to $r_1$.

\item $r_1$ (to $r_2$):  ``I can do so after two steps but I would also like for you ($r_2$) to move out of $l_2$, $l_4$, and $l_5$ after I move out of those places.'' This means that $r_1$ agrees to satisfy the formula sent by $r_2$ but also has some conditions of its own. This can be represented by the formula $\varphi_1 \supset   \varphi_2 = \neg at(r_2, l_2) \wedge \neg at(r_2, l_4) \wedge \neg at(r_2, l_5)$.

\item $r_2$ (to $r_1$): ``that is good; however, do not move through $l_4$ to get out of the area.''

\item etc.

\end{itemize}
The negotiation will continue until either the agent accepts (or refutes) the latest proposal from the other agent. A formal ASP based negotiation framework (e.g.,  the system described by \citeN{sopongsa14a}) could be used for this purpose.

Observe that during a negotiation, none of the robots changes its location or executes any action. After a successful negotiation, each robot has some additional information to take into consideration in its planning. In this example, if the two robots agree after the second proposal by $r_2$, robot $r_1$ agrees to move out of $l_2$, $l_3$, and $l_4$  but should do so without passing by $l_4$; robot $r_2$ knows that he can have $l_2$, $l_3$, and $l_4$ for itself after sometime and also knows that it can stand at $l_4$ until $r_1$ is out of the requested area; etc.   Note, however, that this is not yet sufficient for the two robots to achieve their goals. To do so, they also need to agree on the timing of their moves. For example, $r_1$ can tell $r_2$ that  $l_2$, $l_3$, and $l_4$ will be free after two steps; $r_2$ responds that, if it is the case, then $l_2$, $l_4$, and $l_5$ will be free after 2 steps; etc. This information will help the robots come up with plans for their own goals.

To the best of our knowledge, only a prototype implementation of the approach to interleaving negotiation and planning has been presented \cite{soposa09a}. It is also not implemented distributedly.

\begin{remark}
\begin{enumerate}

\item There are two different ways to enforce the collision-free constraint in the MAPF encoding. One can, for example,
replace \eqref{e-collision} with the rule
\[
\begin{array}{rcl}
& \leftarrow & agent(R), agent(R'), R \ne R' ,  \nonumber \\
& & holds(at(R, V), T), holds(at(R', V'), T), \nonumber \\
& & occ(move(R, V, V'), T), occ(move(R', V', V), T+1) \nonumber
\end{array}
\]

\item  ASP-based solutions for various extensions of the MAPF problems have been discussed by \citeN{ngobsoscye17a} and \citeN{goheba20a}. The encoding proposed by \citeN{goheba20a} is special in that its grounded program has a linear size to the number of agents.
 An ASP-based solution for this problem has been applied in a real-world application \cite{geobscra18a}.
 An environment for experimenting with MAPF has been developed by \citeN{geobotscsangso18a}.
 A preliminary implementation of a MAPF solver on distributed platform can be found in the paper by \citeN{pisotoye19a}.

\item We observe that little attention has been paid to answer set programming based distributed planning. This also holds for the answer set programming based distributed computing platforms. Perhaps the need to attack problems in multi-agent systems will eventually lead to a truly
distributed platform that could push the investigation of using answer set programming in this research direction to the next level. We note that the need for such platform exists and ad-hoc combinations with other programming language have been developed by \citeN{lesopoye15a}.
\end{enumerate}
\end{remark}

\subsection{Context: Planning in Multi-Agent Environments}

Planning in multi-agent environments has been extensively investigated by the multi-agent research community.
There exists a broad literature in this direction which addresses several issues, such as coordination, sharing
of resources, use of shared resources, execution of joint actions, centralized or distributed computation of plans, sharing
of tasks, etc.
Earlier works in multi-agent planning (e.g., see the papers \cite{allzil09a,bradom08a,begiimzi02a,brenner03a,crjoro14a,durfee99a,webotowi03a,weecle09a,golzil04a,gukopa01a,natayopyma03a,nisbra12a,pessav02a,sholey09a,toonsa12a,vlassis07a}) focus on generating plans for multiple agents, coordinating the execution of plans, and does not take into consideration knowledge, beliefs, or privacy of agents.
Work in planning for multiple self-interested agents can be found in the papers \cite{gmydos05a,radogm06a,poubou03a,sondos15a}.


The planning problem in multi-agent environments discussed in this section focuses on the setting in which each agent has its own goal, similar to the setting discussed in earlier work on multi-agent planning.
It should be noted that MAPF has  attracted a lot of attention in recent years due to its widespread applicability such as in warehouse or airtraffic control, leading to the organization of the yearly MAPF workshop at IJCAI and/or ICAPS conferences and several tutorials on the topic.
A good description of this problem can be found in the papers \cite{basvsknokr19a,ststfekomawaliatcokubabo19b}.
Challenges and opportunities in MAPF and its extensions have been described by \citeN{salste20a}.
As with planning, search-based approaches to solving MAPF are frequently used.
Early MAPF solvers, such as the ones described in the papers \cite{gofestshsthosc14a,wagcho15a,shstfest15a,bofestshtobesh15a,courkuxuayko16a,wanbot11a,lunbek11a,wimowi14a},  can compute optimal, boundedly-suboptimal, or suboptimal solutions of MAPF.
\citeN{erkiozsc13a}, \citeN{yulav16a}, and \citeN{sufestbo16a} applied 
answer set programming, mixed-integer programming, and satisfiability testing, respectively, to solve the original MAPF problem.
Suboptimal solutions of MAPF using SAT is discussed by \citeN{sufestbo18a}.
\citeN{surynek19a} presents an SMT-based MAPF solver.

Several extensions of the MAPF problem have been introduced.
\citeN{makoe16a} generalize MAPF to \emph{combined Target Assignment and Path Finding (TAPF),} where agents are partitioned into teams and each team is given a set of targets that they need to reach.
MAPF with deadlines is introduced by \citeN{mawafelikuko18a}. Extensions of the
MAPF problem with  delay probabilities have been described by \citeN{makuko17a}.
The answer set planning implementation by \citeN{ngobsoscye17a} shows that TAPF can be efficiently solved by answer set planning in multi-agent environments.

\citeN{anyaatst19a} investigate MAPF with continuous time, which removes the assumption that transitions between nodes are uniform.
\citeN{barsva19a} present a SAT-based approach to deal with this extension, while
\citeN{surynek19b} describe an SMT-based MAPF solver for MAPF with continuous time and geometric agents.

\citeN{atstfewabazh20a} focus on the issue of unexpected delays of agents and introduce the notion of
$k$-robust MAPF plan, which can still be successfully executed when at most $k$ delays happen.
This paper also studies a probabilistic extension of $k$-robust MAPF plan, called $pk$-robots MAPF plan.

A more realistic version of MAPF, which allows agents to exchange packages and transfer payload, is considered by \citeN{matoshkuko16a}. Discussion of the problems where robots have kinematic constraints can be found in the paper \cite{hokucomaxuayko16a}.

It is worth noting that all of the aforementioned approaches to solving MAPF are centralized. \citeN{pisotoye19a} propose a distributed ASP-based MAPF solver.
In a recent paper, \citeN{goheba21a} present a compact ASP encoding for solving optimal sum-of-cost MAPF that is competitive with other approaches.

\section{Planning with Scheduling and Extensions of ASP}
\label{sec:extension}

Research on applications of ASP planning has also given impulse to, and is intertwined with, work on extensions of ASP. Let us consider the scenario from the following example.

\begin{example}[From the paper by \citeN{balduccini11a}]\label{ezcsp:ex}
In a typical scenario from the domain of industrial printing, orders for the printing of
books or magazines are more or less continuously received by the print shop.
Each order involves the execution of multiple jobs. First, the pages are
\emph{printed} on (possibly different) press sheets. The press sheets are often large enough
to accommodate several (10 to 100) pages, and thus a suitable layout
of the pages on the sheets must be found. Next, the press sheets
are \emph{cut} in smaller parts called \emph{signatures}.
The signatures are then \emph{folded} into booklets whose page size equals the intended
page size of the order. Finally the booklets are \emph{bound} together to form the
book or magazine to be produced. The decision process is made more complex by
the fact that multiple models of devices may be capable of performing a job.
Furthermore, many
decisions have ramifications and inter-dependencies. For example, selecting a large
press sheet would prevent the use of a small press.
The underlying decision-making process is often called \emph{production planning}.
Another set of decisions deals with \emph{scheduling}. Here one needs to
determine \emph{when} the various jobs will be executed using the devices
available in the print shop. Multiple devices of the same
model may be available, thus even competing jobs may be run in parallel.
Conversely, some of the devices can be offline---or go suddenly offline while production is in progress -- and the scheduler must work around that. Typically,
one wants to find a schedule that minimizes the tardiness of the orders while
giving priority to the more important orders.
Since orders are received on a continuous basis, one needs to be able to update
the schedule in an incremental fashion, in a way that causes minimal disruption
to the production, and can satisfy \emph{rush orders}, which need to be
executed quickly and take precedence
over the others. Similarly, the scheduler needs to react to
sudden changes in the print shop, such as a device going offline during production.
\hfill $\Diamond$
\end{example}
This problem involves a combination of planning, configuration (of the devices involved) and scheduling. While ASP can certainly be used to \emph{represent} the problem, computation presents challenges. In particular, the presence of variables with large domains has a tendency to cause a substantial increase in the size of the grounding of ASP programs. Under these conditions, both the grounding process itself and the following solving algorithms may take an unacceptable amount of time and/or memory. Similar challenges have been encountered in the ASP encoding of planning problems with large number of actions and steps (see, e.g., the discussion by  \citeN{sonpon07a}).

Various extensions of ASP have been proposed over time to overcome this challenge. Some approaches, e.g., in response to large planning problems, rely on the avoidance of grounding, as illustrated in systems with lazy grounding (see, e.g., the papers by \citeN{padoporo09a}, \citeN{cadebrsr15a}, and \citeN{tawefr19a}) or on the use of top-down execution models (see, e.g., the papers by \citeN{boposo08a} and \citeN{margup12a}). At the core of the attempts focused on combination of planning and scheduling is the integration of ASP with techniques from constraint solving; this approach enables the effective ability to handle variables with large domains (especially numerical) efficiently.

\citeN{elposo04a} provided an initial exploration of the combination of answer set programming with constraint logic programming, mostly focused on supporting the introduction of aggregates in answer set programming.
\citeN{baboge05a} were among the first to propose a methodology for achieving such an integration as a way to extend the capabilities of the answer set programming framework. In their approach, the syntax and semantics of ASP is extended to enable the encoding of numerical constraints within the ASP syntax.
\citeN{megezh08a} and \citeN{geossc09a} proposed solvers that support variants of ASP defined along the lines of the approach by \citeN{baboge05a}, and specific ASP and constraint solvers are modified and integrated. In particular, the approach discussed in the \emph{clingcon} systems \cite{ostsch12a,bakaossc16a} explores the integration of constraint solving techniques with techniques like clause learning and back-jumping.


Experimental results show that these approaches lead to  increased scalability, enabling the efficient resolution
of planning domains of larger size. Later research extends and generalizes the language  \cite{barlee13a}, increasing
 the efficiency of the resolution algorithms; these extensions have been
  eventually integrated in the mainstream \clingo\ solver \cite{gekakaosscwa16a}.
Notably, its extension with difference constraints, viz.\ \clingoM{dl}, is operationally used by Swiss Railway for
routing and scheduling train networks~\cite{abjoossctowa19a}.

\begin{figure}[htbp]
\includegraphics[clip=true,trim=0 170 0 0,width=1\textwidth]{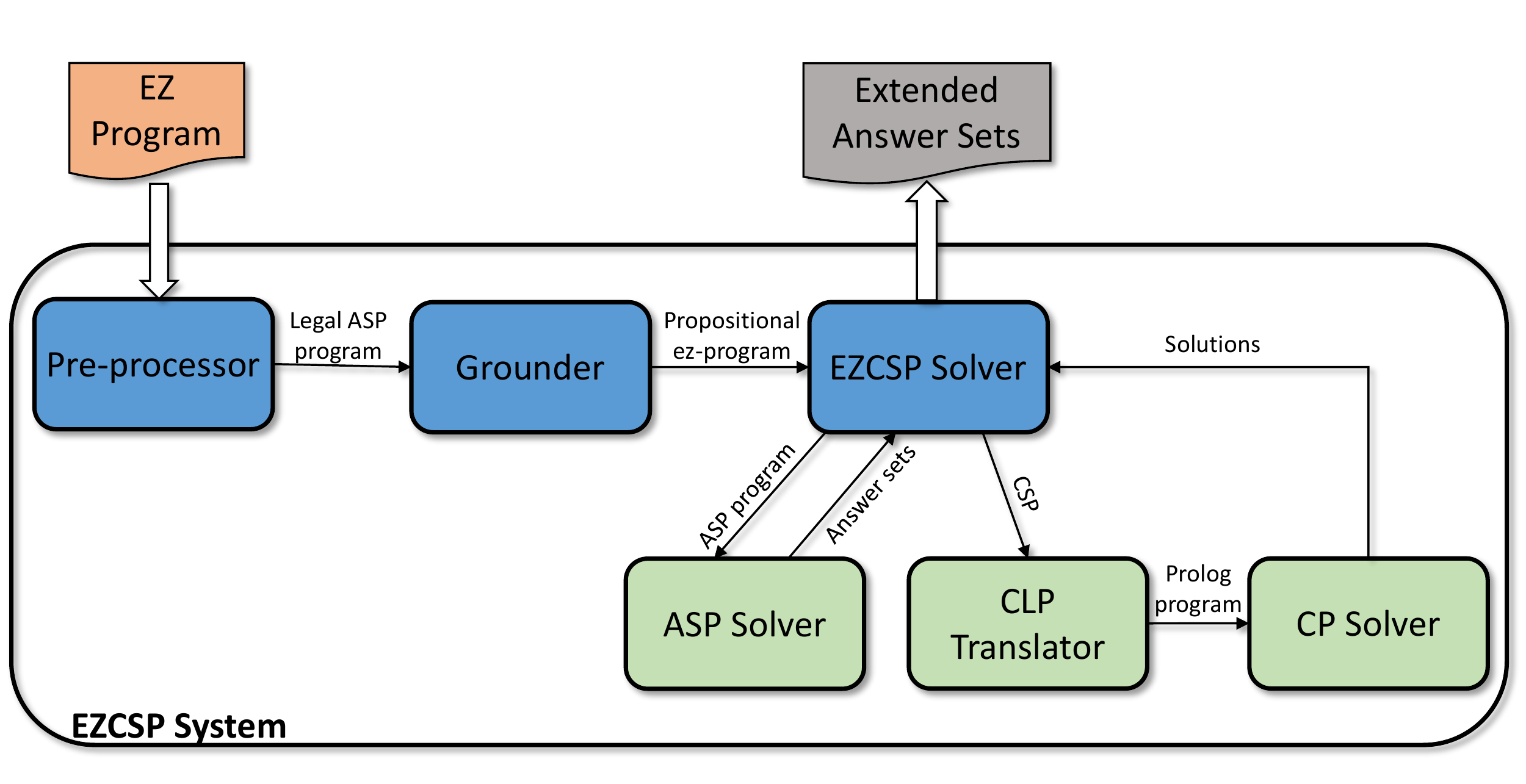}
\caption{Architecture of the \textsc{EZCSP} solver \protect\cite{balduccini11a}}
\label{fig:ezcsp-arch}
\end{figure}

All of these approaches rely on a \emph{clear-box} architecture \cite{ballie13a}, i.e., an architecture where the ASP components of the algorithm and its constraint solving components are tightly integrated and modified specifically to interact with each other. A different approach is proposed by \citeN{balduccini09a} and later extensions. In that line of research, the goal is to enable the reuse, without modifications, of existing ASP and constraint solving algorithms. The intuition is that such an arrangement  allows one to employ the best solvers available and  makes it easy to analyze the performance of different combinations of solvers on a given task. This enabled researchers to propose a \emph{black-box} architecture (and, later, a more advanced \emph{gray-box} architecture), where the ASP and the constraint solving components are unaware of each other and are connected only by a thin ``upper layer'' of the architecture, which is  responsible for exchanging data between the components and triggering their execution. The architecture proposed by \citeN{ballie17a} is illustrated in Figure \ref{fig:ezcsp-arch}. Intuitively, answer sets are computed first by an off-the-shelf ASP solver. Special atoms are gathered by the ``upper layer'' and translated into a constraint satisfaction problem, which is solved by an off-the-shelf constraint solver. Solutions to the overall problem correspond to pairs formed by an answer set and a solution to the constraint satisfaction problem extracted from that answer set.

This approach  features an embedding of constraint solving constructs directly within the ASP language---that is, without the need to extend the syntax and semantics of ASP---by means of pre-interpreted relations known to the ``upper layer'' of the architecture (and some syntactic sugars for increased ease of formalization). For instance, constraint variables for the start time of jobs from Example \ref{ezcsp:ex} can be declared in the language described in the paper \cite{balduccini11a} by means of the rule
\begin{equation}\label{ezcsp:cspvar}
cspvar(st(D,J),0,MT) \leftarrow
job(J),\ job\_device(J,D),
max\_time(MT)
\end{equation}
where $cspvar$ is a special relation that the ``upper layer'' knows how to translate into a variable declaration for a numerical constraint solver. Similarly, the effect on start times of precedences between jobs can be encoded by the ASP rule
\begin{equation}\label{ezcsp:required}
\begin{array}{l}
required(st(D2,J2) \geq st(D1,J1) + Len1) \leftarrow \\
\hspace*{1in} job(J1),\ job(J2), \
job\_device(J1,D1),\ job\_device(J2,D2), \\
\hspace*{1in} precedes(J1,J2),\ 
job\_len(J1,Len1)
\end{array}
\end{equation}
where ``$\geq$'' is a syntactic sugar for the predicate $at\_least$, written in the infix notation, and it is replaced by a pre-interpreted function symbol during pre-processing. As before, the ``upper layer'' is aware of the relation $required$ and translates the corresponding atoms to numerical constraints. A problem involving a combination of planning and scheduling such as that described in Example \ref{ezcsp:ex} can be elegantly and efficiently solved by extending a planning problem $\lan D, \Gamma, \Delta \ran$ with statements such as (\ref{ezcsp:cspvar}) and (\ref{ezcsp:required}).

Later research on this topic \cite{ballie13a,ballie17a} uncovered an interesting result: contrary to what one might expect, there is no clear winner in the performance comparison between black-box and white-box architectures (and gray-box as well); different classes of problems are more efficiently solved by a different architecture.
Furthermore, \citeN{bamamale17a} showed that the EZCSP architecture can be used in planning with PDDL+ domains.


\section{Conclusions and Future Directions}\label{sec:discussion}


This paper surveys the progress made over the  last 20+ years in the area of
\emph{answer set planning.} It focuses on the  encoding in answer set programming of different
classes of planning problems: when the initial state is complete, incomplete, and with or without sensing actions. In addition, the paper shows that answer set planning can reach the level of scalability and efficiency of state-of-the-art specialized planners, if useful information which can be exploited to guide the search process in planning, such as heuristics, is provided to the answer set solver. The paper also reviews some of the main research topics related to planning, such as planning with preferences, diagnosis, planning in multi-agent environments, and planning integrated with scheduling. We note that
research related to answer set planning has been successfully applied in different application domains, often in combination with other types of reasoning, such as planning for the shuttle spacecraft by \citeN{nobagewaba01a}, planning and scheduling \cite{balduccini11a}, robotics \cite{akererpa11a}, scheduling \cite{abjoossctowa19a,dogakhmapo19a,geobscra18a}, and multi-agent path findings \cite{goheba21a,ngobsoscye17a}. Section~\ref{sec:extension} also discusses the potential impacts that the use of answer set planning in real-world applications can have on the development of answer set programming.

In spite of this extensive body of research, there are still several challenges for answer set planning.

\paragraph{Performance of ASP Planning:} We note that extensive experimental comparisons with state-of-the-art compatible planning systems have been conducted.
For example, \citeN{gekaotroscwa13a} experimented with classical planning, \citeN{eifalepfpo03b} and \citeN{tusogemo11a} worked with incomplete information and non-deterministic actions, and \citeN{tusoba07a} with planning with sensing actions.
The detailed comparisons can be found in the aforementioned papers and several other references that have been discussed throughout the paper.
These comparisons demonstrate that ASP-planning is competitive with other approaches to planning, such as heuristic based planning or SAT-based planning.

The flexibility and expressiveness of answer set programming provide a simple way for answer set planning to exploit various forms of domain knowledge.
To the best of our knowledge, no heuristic planning system can take advantages of all  well-known types of domain knowledge, such as hierarchical structure, temporal knowledge, and procedural knowledge, whist they can be easily integrated into a single answer set planning system, as demonstrated by \citeN{sobanamc03a}.

It is important to note that the performance of answer set planning systems depends heavily on the performance of the answer set solvers used in computing the solutions. 
As such, it is expected that these answer set  planing systems can benefit from the advancements made by the ASP community.
On the other hand, this hand-off approach also gives rise to  limitations of ASP-based planning systems, such as scalability, heuristics, and ability to work with numeric values.
It is worth noting that heuristics can be specified for guiding the answer set solver (e.g., as done by \citeN{gekaotroscwa13a} in planning) and there are considerable efforts in integrating answer set solvers and constraint solvers (see Section~\ref{sec:extension}).
However, there exists no answer set planning system that works with PDDL+, similarly to the system SMTPlan by
\citeN{camaze20a} which employs SMT planning for hybrid systems described in PDDL+.
An ASP-based planning system for PDDL+ is proposed by \citeN{bamamale17a}. Yet, this system cannot deal with all features of PDDL+ as SMTPlan.
The issue of numeric constraints is also related to the next challenge.

%

\paragraph{Probabilistic planning:} This research topic is the objective of intensive research  within the automated planning community. Similarly to other planning paradigms, competitions among different probabilistic planning systems are  organized within ICAPS (International Conference on Automated Planning and Scheduling, e.g., \url{https://ipc2018-probabilistic.bitbucket.io/})  and attract several research groups from academia and industry.

Probabilistic planning is concerned with identifying an optimal policy for an agent in a system specified by a
\emph{Markov decision problem (MDP)} or a \emph{Partial observable MDP (POMDP).} While algorithms for computing an optimal policy are readily available (e.g., value iteration algorithm by \citeN{bellman57a}, topological value iteration by \citeN{damawego11a}, ILAO* by \citeN{hanzil01a}, LRTPD by \cite{bongef03a}, UCB by \citeN{kocsze06a}, as well
as the work of \citeN{kalica98a}; the interested readers is also referred to the survey by \citeN{shpika13a}), scalability and efficiency remain significant issues in this research area.

Computing MDP and POMPD in logic programming will be a significant challenge for ASP, due to the fact that answer set solvers are not developed to easily operate with real numbers.
In addition, the exponential number of states of a MDP or POMPD require a representation language suitable for use with ASP.
This challenge can be addressed using probabilistic action languages, such as the ones proposed by \citeN{banatu02a} or by \citeN{wanlee19a}.
On the other hand, working with real numbers means that the \emph{grounding-then-solving} method
to compute answer sets is no longer adequate.
First of all, the presence of real numbers implies that the grounding process might not terminate.
While discretization can be used to alleviate the grounding problem, it could increase the size of the  grounded program significantly, creating problems (e.g., lack of memory) for the solving process.
A potential approach to address these challenges is to integrate constraint solvers into answer set solvers, creating \emph{hybrid systems}  that can effectively deal with numeric constraints. Research in this direction has been
summarized in  Section~\ref{sec:extension}.
The recent implementation by \citeN{abjoossctowa19a} demonstrates that answer set programming based system can work effectively with a huge number of numeric constraints.

%

\paragraph{Epistemic planning:}
In recent years, epistemic multi-agent planning (EMP) has gained significant interest within the planning community.
\citeN{lopawi11a} propose a general epistemic planning framework.
Complexity of EMP has been studied in the papers \cite{aucbol13a,bojesc15a,chmasc16a}.
Studies of EMP can be found in several papers \cite{boland11a,enbomane17a,hoewoo02a,hufawali17a,lopawi11a,eijck04a}
and
many planners have been developed  \cite{bufadopo20a,fabudopo20a,lefasopo18a,mubefemcmipeso15a,komgef15a,komgef17a,wayafalixu15a}.
With the exception of the planner developed by \citeN{bufadopo20a}, which employs answer set programming,
the majority of the proposed systems  are heuristic search based planners.
Some EMP planners, such as those proposed by \citeN{mubefemcmipeso15a}, \citeN{komgef15a} and \citeyear{komgef17a},  translate an EMP problem into a classical planning problem and use classical planners to find solutions.

A multi-agent planning problem of this type is different from the planning problems
discussed in Section~\ref{sec:mas}, in that it considers the knowledge and beliefs of agents, and explores the use of actions that manipulate such knowledge and beliefs. This is necessary for planning in non-collaborative
and competitive environments. The difficulty in this task lies in that the result of the execution of an action (by an agent or a group of agents) will change the state of the world and the state of knowledge and belief of other agents. Inevitably, some agents may have false beliefs about the world. Semantically, the transitions from the state of affair, that includes the state of the world and the state of knowledge and beliefs of the agents, to another state of affair could be modeled by transitions between Kripke structures (see, e.g., the books \cite{fahamova95a,dihoko07a}).
A Kripke structure consists of a set of worlds and a set of binary accessibility relations over the worlds. A practical challenge is related to  the size of the Kripke structures, in terms of the number of worlds---as this can  double after the execution of each action. Intuitively, this requires the ability to generate new terms in the answer set solvers during resolution. Multi-shot solvers (see, e.g., the paper \cite{gekakasc17a}) could provide a good platform for epistemic planning.
Preliminary encouraging results on the use of answer set programming in this context have been recently
presented by \citeN{bufadopo20a}.

\paragraph{Explainable Planning (XAIP):} This is yet another problem that has only recently been investigated  but attracted considerable attention from the planning community, leading to the organization of a a yearly workshop on XAIP associated with ICAPS (e.g.,
\url{https://icaps20subpages.icaps-conference.org/workshops/xaip/}).
Several questions for XAIP are discussed by \citeN{foloma17a}.
In XAIP, a human questions the planner (a robot's planning system) about its proposed solution.
The focus is on explaining \emph{why} a planner makes a certain decision.
For example, why an action is included (or not included) in the plan? Why is the proposed plan optimal?
Why can a goal not be achieved?
Why should  (or should not) the human consider replanning?

\citeN{chsrzhka17a}, for example, describe model reconciliation problem (MRP) and propose methods for solving it.
In a MRP,  the human and the robot have their own planning problems.
The goal in both problems are the same, but the action specifications and the initial states might be different.
The robot generates an optimal plan and informs the human of its plan.
The human declares that it is not an optimal plan according to their planning problem specification.
The robot, which is aware of the human's problem specification, needs to present to the human the reasons why its plan is optimal and what is wrong in the human's problem specification.
Often, the answer is  in the form of a collection of actions, actions' preconditions and effects, and literals from the initial states that should be added to, or removed from, the human's problem specification so that the robot's plan will be an optimal plan of the updated specification.
While explanations have been extensively investigated by the logic programming community, explainable planning using answer set programming has been investigated only recently by  \citeN{ngstsoye20a}.


\subsection*{Acknowledgement}

Tran Son and Enrico Pontelli have been partially supported by NSF grants 1914635, 1833630, 1757207, and 1812628.
Tran Son's and Marcello Balduccini's contribution was made possible in part through the help and support of NIST via cooperative agreement 70NANB21H167.
Torsten Schaub was supported by DFG grant SCHA~550/15, Germany.

\bibliographystyle{acmtrans}

\bibliography{bibexport}

\end{document}